%% file: neurips_2024.tex
\documentclass{article}

\usepackage[preprint]{neurips_2023}

\usepackage[utf8]{inputenc} 
\usepackage[T1]{fontenc}    
\usepackage{hyperref}       
\usepackage{url}            
\usepackage{booktabs}       
\usepackage{amsfonts}       
\usepackage{nicefrac}       
\usepackage{microtype}      
\usepackage{xcolor}         
\usepackage{amsmath,amsfonts,amsthm,amssymb,bm, verbatim,dsfont,mathtools}
\usepackage{color,graphicx,appendix}
\usepackage{tikz}
\usepackage{caption}
\usepackage{amssymb}
\usepackage{pifont}
\usepackage{booktabs}
\usepackage{subcaption}
\usepackage{longtable}
\usepackage{array}
\usepackage{geometry}
\usepackage{booktabs}
\usepackage{subcaption}
\usepackage{graphicx}
\definecolor{mylightblue}{RGB}{173,216,230} 

\title{CHESS: Contextual Harnessing for Efficient SQL Synthesis}

\author{
  Shayan Talaei \\
  Stanford University\\
  \texttt{stalaei@stanford.edu} \\
  \And
  Mohammadreza Pourreza \\
  University of Alberta \\
  \texttt{pourreza@ualberta.ca} \\
  \AND
  Yu-Chen Chang \\
  Stanford University \\
  \texttt{cjohnny@stanford.edu} \\
  \And
  Azalia Mirhoseini\thanks{Equal senior authorship} \\
  Stanford University \\
  \texttt{azalia@stanford.edu} \\
  \And
  Amin Saberi\footnotemark[1] \\
  Stanford University \\
  \texttt{saberi@stanford.edu} \\
}

\begin{document}

\maketitle

\begin{abstract}
Utilizing large language models (LLMs) for transforming natural language questions into SQL queries (text-to-SQL) is a promising yet challenging approach, particularly when applied to real-world databases with complex and extensive schemas. In particular, effectively incorporating data catalogs and database values for SQL generation remains an obstacle, leading to suboptimal solutions. We address this problem by proposing a new pipeline that effectively retrieves relevant data and context, selects an efficient schema, and synthesizes correct and efficient SQL queries. To increase retrieval precision, our pipeline introduces a hierarchical retrieval method leveraging model-generated keywords, locality-sensitive hashing indexing, and vector databases. Additionally, we have developed an adaptive schema pruning technique that adjusts based on the complexity of the problem and the model's context size. Our approach generalizes to both frontier proprietary models like GPT-4 and open-source models such as Llama-3-70B. Through a series of ablation studies, we demonstrate the effectiveness of each component of our pipeline and its impact on the end-to-end performance. Our method achieves new state-of-the-art performance on the cross-domain challenging BIRD dataset.
\end{abstract}

\section{Introduction}
\label{introduction}


Translating natural language questions into database queries, or text-to-SQL, is a long-standing research problem. This issue has been exacerbated in recent years due to the growing complexity of databases, driven by the increasing sizes of schemas (sets of columns and tables), values (content), and catalogs (metadata describing schemas and values) stored within them. Even the largest proprietary models, such as GPT-4, lag significantly behind human performance on text-to-SQL benchmarks, with a notable accuracy gap of 30\% \citep{li2024can}. Beyond the complexity of writing SQL queries, this substantial gap is primarily caused by the need to effectively retrieve and integrate multiple sources of information, including database values, catalogs, and schema, each in different formats, which complicates the process.



\begin{figure}[t]
    \centering
    \includegraphics[width=0.99\textwidth]{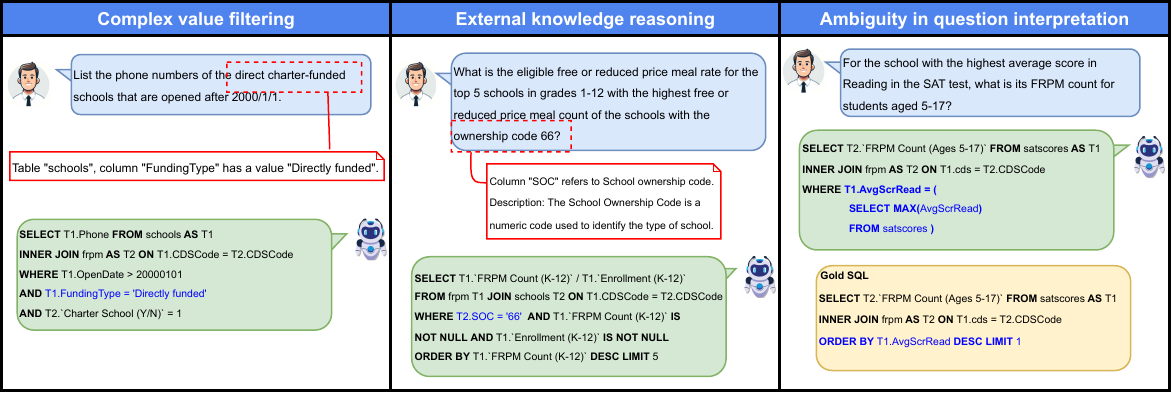}
    \caption{Example of challenges in text-to-SQL translation. 1) Questions passed by the users might not have the exact database value. 2) Column names might not be a good representation of what they store so using database catalogs is an essential part of text-to-SQL translation. 3) For a given question there are multiple ways of writing a correct SQL answer.}
    \label{fig:intro_examples}
\end{figure}

In Figure \ref{fig:intro_examples}, we show some of the challenges facing modern text-to-SQL systems. For instance, users' questions might not directly match the stored values in the database, making it crucial to accurately identify the value format for effective SQL query formulation. Additionally, real-world database schemas often contain ambiguous column names, table names, and messy data, complicating the SQL translation process and necessitating a robust retrieval system to identify relevant information. Moreover, there are typically multiple valid SQL queries that could answer the same question. For example, for the question illustrated on the right side of the Figure \ref{fig:intro_examples}, one might use 'ORDER BY' and 'LIMIT 1' to find the highest average score, while another approach could involve a subquery with the 'MAX()' function, potentially leading to different outputs.

Earlier work in the area \citep{pourreza2024din, wang2023mac, qi2022rasat, rajkumar2022evaluating, li2024pet} has generally limited the context for SQL generation to table structures, column definitions, and sample rows. However, in production-level databases, the database catalog and database values constitute a rich source of information that is crucial for generating accurate SQL queries. 

We introduce ``CHESS: contextual harnessing for efficient SQL
synthesis", an end-to-end text-to-SQL system that targets real-world and complex databases. CHESS introduces a scalable and effective LLM-based pipeline for SQL generation that consists of three main components: entity and context retrieval, schema selection, and SQL generation.


For entity and context retrieval, we present scalable and efficient methods using locality-sensitive hashing to retrieve database values from millions of rows, leveraging keyword detection, and vector databases to extract contextual information from database catalogs. Our approach utilizes both semantic and syntactic similarities between the database content and the user's query to enhance SQL prediction accuracy. 
In the schema selection phase, we utilize the retrieved information to narrow down the initial schema with potentially hundreds of columns to an efficient set of columns, usually less than ten. Throughout this step, we extract a minimal yet sufficient subset of the database schema. Finally, the extracted database schema is passed to a query generation module, which uses our fine-tuned SQL generator model combined with a revision step to effectively generate a SQL query.

Through a series of ablation studies \ref{ablations}, we demonstrate the critical role of each module in the pipeline in guiding LLMs to generate accurate SQL queries. Specifically,  our entity and context retrieval module contributes substantially to performance, as evidenced by a 5\% accuracy improvement.



At the time of submission, CHESS ranks first among all disclosed methodologies on BIRD \citep{li2024can}, with a 65\% and 66.69\% execution accuracy on the development and test set respectively. To our knowledge, BIRD is the most challenging real-world text-to-SQL benchmark that is  publicly available. It features more than 12000 unique question-SQL pairs, spanning 37 professional fields, including healthcare, education, blockchain, and sports, and covering 95 large databases with a combined size of 33.4 GB. An active leaderboard for BIRD is maintained by a third party (with a private test set), which we used to evaluate the performance of CHESS. CHESS also ranks second among all methods, with a marginal gap to the propriety (and undisclosed) approach that currently has a test set accuracy of 67.86\% on BIRD's leaderboard.

We also provide an end-to-end open-source version of CHESS that obtains the best performance among other open-source baselines on the BIRD development set, with an execution accuracy of 61.5\%. The use of open-source models for text-to-SQL is crucial, particularly when databases may contain private information that should not be shared with third-party LLM providers. 
All of the codes to reproduce the results reported in this paper are available on Anonymous Github repository
\footnote{\url{https://github.com/ShayanTalaei/CHESS}}.


Concretely, our contributions are as follows:
\begin{itemize}
    \item Breaking down the Text-to-SQL task into a 3-staged pipeline, including entity and context retrieval, schema selection, and query generation
    \item A scalable hierarchical retrieval approach for extracting the important entities and contexts 
    \item An efficient three-staged schema pruning protocol, consisting of individual column filtering, table selection, and a final column selection for extracting a minimally sufficient schema
    \item A fine-tuned open-source DeepSeek-33B Coder model with a novel training dataset construction approach with noise injection to mitigate error propagation
    \item A high-performing end-to-end open-source pipeline ensuring the privacy of the information
    \item Setting new state-of-the-art results on the   BIRD dataset among known methodologies
\end{itemize}

\section{Related Work}
Generating accurate SQL queries from natural language questions, known as text-to-SQL is an active area of research within both the natural language processing (NLP) and database communities. Early efforts by the database community approach the problem through custom templates \citep{zelle1996learning} marked initial advancements in this field, albeit at the expense of significant manual effort. Recently, text-to-SQL methodologies have increasingly incorporated transformer-based models, particularly sequence-to-sequence architectures \citep{vaswani2017attention, sutskever2014sequence}. These sequence-to-sequence models, capable of end-to-end training, are particularly well-suited for tasks requiring the generation of one sequence from another, such as translation, summarization, and text-to-SQL \citep{qin2022survey}.

Initial sequence-to-sequence models, such as IRNet \citep{guo2019towards}, utilized a bidirectional LSTM neural architecture to encode the query and employed self-attention to encode the database schema representation. To enhance the integration of schema information and capture its relationship with the question, models like RAT-SQL \citep{wang2019rat} and RASAT \citep{qi2022rasat} have employed relation-aware self-attention mechanisms. Additionally, SADGA \citep{cai2021sadga} and LGESQL \citep{cao2021lgesql} have adopted graph neural networks to represent the relational structures between the database schema and the queries. Although sequence-to-sequence models have improved performance, further advancements are necessary to bridge the gap with human performance. For example, none of the above techniques achieve an execution accuracy of more than 80$\%$ on the Spider hold-out test set \citep{yu2018spider}.

Alongside the widespread adoption of LLMs across various NLP domains, the text-to-SQL field has similarly benefited from recent methodological innovations with LLMs to enhance performance. Early approaches \citep{rajkumar2022evaluating}, leveraged the zero-shot in-context learning capabilities of LLMs for SQL generation. Building on this, subsequent models including DIN-SQL \citep{pourreza2024din}, DAIL-SQL \citep{gao2023text}, MAC-SQL \citep{wang2023mac}, and C3 \citep{dong2023c3} have enhanced LLM performance through task decomposition and techniques such as Chains of Thought (CoT) \citep{wei2022chain}, self-consistency \citep{wang2022self}, and least-to-most prompting \citep{zhou2022least}. In addition to in-context learning, proposals in DAIL-SQL \citep{gao2023text}, DTS-SQL \citep{pourreza2024dts}, and CodeS \citep{li2024codes} have sought to elevate the capabilities of open-source LLMs through supervised fine-tuning, aiming to rival or exceed their larger, proprietary counterparts. However, the most notable performance gains have been observed in proprietary LLMs utilizing in-context learning methods \citep{li2024can}. Unlike previous efforts, this paper introduces a hybrid approach that combines both in-context learning and supervised fine-tuning to further enhance performance. Moreover, we propose novel methods to integrate contextual data such as database values and database catalog into the text-to-SQL pipeline, leveraging a rich yet often overlooked source of information.

Independently, but concurrent with our work, MCS-SQL \citep{lee2024mcs} introduced a method that relies on using multiple prompts and sampling several responses from LLMs to mitigate sensitivity to the arrangement of tables, columns, and few-shot in-context samples. They also devised a technique to filter out irrelevant table and column names. However, unlike our approach, they do not emphasize retrieving pertinent information from the database catalog and database values. 
Instead, they rely on using a large number of samples from LLMs to reduce prompt sensitivity for accuracy improvement. 

\section{Methodology}
\label{sec:methodology}



The text-to-SQL task is a challenging learning problem, requiring the model to translate a natural language question and database schema into a valid SQL query. The input consists of the text of the question, the database schema defining table structures and column types, and the database instance containing the actual or samples of the table content. Further, the input may include a database catalog with metadata, allowing free-form textual attribute values. Key challenges include noisy database content, the need to implicitly capture semantic correspondences between input elements, and the compositional nature of mapping language to SQL's formal query structure.

\begin{figure}[h]
    \centering
    \includegraphics[width=0.8\textwidth]{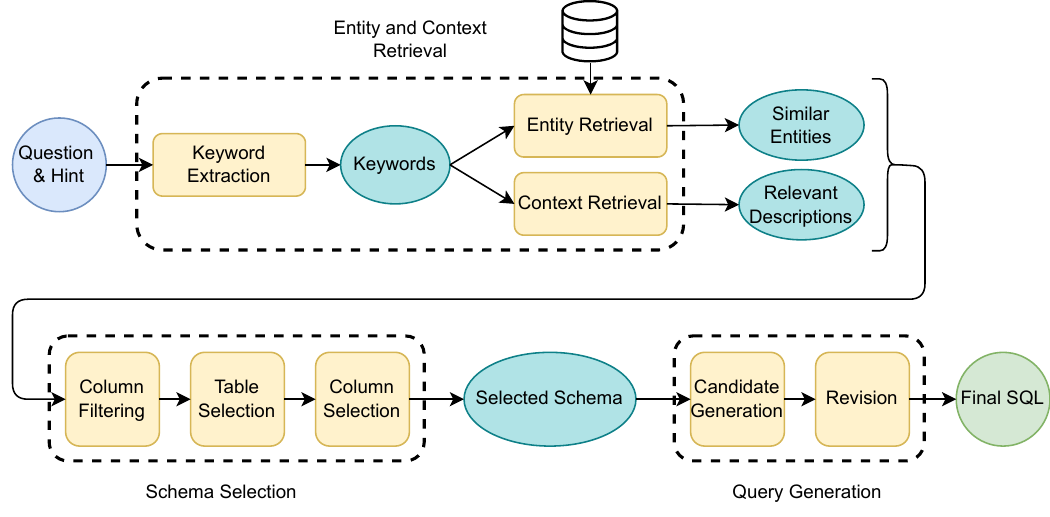}
    \caption{Our  pipeline with   modules for entity and context retrieval, schema selection, and query generation.}
    \label{fig:full_pipeline}
\end{figure}

"CHESS: Contextual Harnessing for Efficient SQL Synthesis" is an end-to-end text-to-SQL system designed for complex, real-world databases. It introduces a scalable and effective LLM-based pipeline consisting of three main components: entity and context retrieval, schema selection, and SQL generation. The entity and context retrieval component \ref{subsec:relevant_information_retrieval} uses keyword selection, locality-sensitive hashing,  and vector databases to efficiently retrieve relevant database values and contextual information from large databases. The schema selection phase \ref{subsec:schema_selection} then narrows down the initial schema to a minimal yet sufficient subset of columns. Finally, the extracted schema is passed to a query generation module \ref{subsec:query_generation} that utilizes a fine-tuned SQL generator model and a revision step to generate an accurate SQL query. The overall pipeline is demonstrated in Figure \ref{fig:full_pipeline}. 

Finally, some of the implementation details of our method, including how we preprocess database values and data catalogs to expedite retrieval during the pipeline execution, are provided in Appendix \ref{sec:implementation_details}. Detailed execution traces of our pipeline are presented in Appendix \ref{sec:execution_trace}.



A key feature of our SQL query generation approach is the integration of multiple sources of information, including database catalogs, values, and schemas. However, these datasets with schemas containing hundreds of columns, each with detailed descriptions, and potentially millions of rows are often very large. Passing all this information to an LLM is often impractical due to limited context windows. Even if feasible, it can negatively impact the LLM's reasoning capabilities, as demonstrated in \citep{hsieh2024ruler} and our own ablation studies in Section \ref{complexity_analysis}. Our pipeline addresses this challenge by providing the LLM with minimal yet sufficient information necessary for each task. Maintaining \emph{minimal sufficiency} is a key feature in all modules of the pipeline. Most crucially, during the SQL generation phase, we try to identify and pass to the model only the columns that are needed for the generation of the SQL query.

\vspace{-0.3em}
\subsection{Entity and Context Retrieval}
\label{subsec:relevant_information_retrieval}
\vspace{-0.3em}

The first module in the pipeline  identifies the relevant information in the input, including the entities referred to in the question and the contextual information provided about them in the database schema. This is done in three steps. 
\vspace{-0.7em}
\hypertarget{subsubsec:keyword_extraction}{\paragraph{Keyword Extraction.}}
To search for similar values in the database and schema descriptions, we first need to extract the main keywords from the natural-language question. Our approach to this problem is to prompt the model with few-shot examples of the task and the question, asking it to identify and extract
keywords, keyphrases, and named entities. 
\vspace{-0.7em}
\hypertarget{subsubsec:entity_retrieval}{\paragraph{Entity Retrieval.}}
From the list of keywords extracted from the question, some may correspond to entities present in the database values. In this step, we search for similar values in the database and return the most relevant ones, along with their corresponding columns, for each keyword. As illustrated in Figure~\ref{fig:intro_examples}, searching for exact matches of the keywords cannot handle variations or typos, necessitating a more flexible search method.
To measure the syntactic similarity between the keywords and the database values, we use the edit distance similarity metric. Additionally, to make the retrieval process more efficient, we propose a hierarchical retrieval strategy based on \hyperlink{subsubsec:lsh_indexing}{Locality Sensitive Hashing (LSH)} and semantic (embedding) similarity measures, which we explain in detail in Appendix \ref{sec:implementation_details}. This approach allows us to efficiently retrieve values that exhibit a high degree of both syntactic and semantic similarity to the keywords.


\vspace{-0.7em}
\hypertarget{subsubsec:context_retrieval}{\paragraph{Context Retrieval.}} In addition to the values, database catalogs explaining the schema may be available. For instance, each column may have a description, an extended column name (in the case of abbreviations), and a value description. As shown in Figure~\ref{fig:intro_examples}, this information can be useful, and not providing it to the model can lead to suboptimal performance. As explained before, in retrieving this context, we aim to identify only the minimally sufficient or the most relevant information. This is done by retrieving the most similar descriptions to the extracted keywords, measured by a semantic (embedding) similarity metric when querying the \hyperlink{subsubsec:vector_database}{vector database of descriptions} created during the preprocessing step.



\vspace{-0.3em}
\subsection{Schema Selection}
\vspace{-0.3em}
\label{subsec:schema_selection}
Our goal in this step is to narrow down the schema to include only the necessary tables and columns needed for generating the SQL query. \hypertarget{efficient_schema}{We refer to this optimized set of necessary tables and columns as the \emph{efficient schema}}. Achieving an efficient schema leads to better performance in SQL query generation by excluding irrelevant information.
We use recall and precision metrics to determine whether we have selected the correct tables and columns using the correct SQL query as the ground truth. The results are presented in Table \ref{table:recall_precision} along with a detailed example of the schema selection process in Appendix \ref{sec:schema_selection_example}. The prompts for all of the sub-modules in schema selection are augmented with chain-of-though prompting \cite{wei2022chain} to improve the reasoning ability of LLMs which is essential for this task.


\vspace{-0.7em}
\hypertarget{subsubsec:column_filtering}{\paragraph{Individual Column Filtering.}}
A database can contain hundreds of columns, many of which may be semantically irrelevant to the question. Looking for an efficient schema, we aim to filter out the irrelevant columns and pass only the most relevant ones to the \hyperlink{subsubsec:table_selection}{table selection} step.
To accomplish this, we treat the relevance of each column to the question as a binary classification task for the model, essentially asking the LLM if the column may be relevant to the question. This step is useful only for removing the irrelevant columns that are obvious, but evaluating the relevance of a column in isolation is not always possible.  We address this limitation in the subsequent steps, \hyperlink{subsubsec:table_selection}{table selection} and \hyperlink{subsubsec:column_selection}{column selection}, in which we give the model a more global view of the schema.

\vspace{-0.7em}
\hypertarget{subsubsec:table_selection}{\paragraph{Table Selection.}}
After filtering out irrelevant columns, we proceed to select the tables essential for generating the SQL query. In this step, we present the model with the filtered schema from the previous step and ask it to assess the relevance of each table, selecting only those necessary for the SQL query.


\vspace{-0.7em}
\hypertarget{subsubsec:column_selection}{\paragraph{Final Column Selection.}}
In the final step of schema selection, we aim to reduce the schema to the minimal set of columns necessary for generating the SQL query. We prompt the model to evaluate the necessity of each column in the filtered tables. This step includes a chain-of-thought explanation of why each column is needed, followed by the selection of the required columns.

\vspace{-0.3em}
\subsection{Query Generation}
\label{subsec:query_generation}
\vspace{-0.3em}

At this point, we have selected an efficient schema augmented by the relevant context, containing all the necessary information to craft a SQL query that answers the question. In the following steps, we first write a candidate SQL query and then revise it to fix potential semantic and syntactic errors.

\vspace{-0.7em}
\hypertarget{subsubsec:candidate_generation}{\paragraph{Candidate Generation.}}
After reducing the schema to the minimal set of tables and columns, we prompt the model to generate an SQL query that answers the question. In the prompt, we provide the minimal schema obtained from the previous steps, along with the relevant values and descriptions retrieved in the first step of the pipeline. With this information, the model generates a candidate SQL query.

\vspace{-0.7em}
\hypertarget{subsubsec:revision}{\paragraph{Revision.}}
In the final step of the pipeline, we aim to fix potential logical and syntactic errors in the candidate SQL query. We provide the model with the database schema, the question, the generated candidate SQL query, and its execution result. The model is then asked to evaluate the correctness of the SQL query and revise it if necessary.

To assist the model in identifying and correcting mistakes, we give it a set of rules following \citep{bai2022constitutional}. In some cases, there may be multiple ways to correct the candidate SQL query. Even with zero-temperature sampling, the model might output different corrections across multiple samplings. To reduce the noise in the model's output, we use self-consistency to select the SQL query that appears most consistently across three samples.

\vspace{-0.3em}
\subsection{Preprocessing}
\label{subsec:preprocessing}
\vspace{-0.3em}

To facilitate the information retrieval process outlined in \ref{subsec:relevant_information_retrieval} efficiently, we preprocess database values and catalogs before executing the pipeline. For database values, we conduct a syntactic search by creating a Locality Sensitive Hashing (LSH) index, as described in \hyperlink{subsubsec:entity_retrieval}{entity retrieval}. For database catalogs, which contain longer texts requiring semantic understanding, we use a \hyperlink{subsubsec:vector_database}{vector database} retrieval method to measure semantic similarity.

\vspace{-0.7em}
\hypertarget{subsubsec:lsh_indexing}{\paragraph{Locality Sensitive Hashing Indexing of Values.}}

To optimize the entity retrieval step, we employ a method capable of efficiently searching through large databases, which may contain millions of rows, to retrieve the most similar values. This step doesn't require perfect accuracy but should retrieve a reasonably small set of similar values, such as a hundred elements.
Locality Sensitive Hashing (LSH) is an effective technique for approximate nearest-neighbor searches. It allows us to retrieve database values that are most similar to a given keyword. During preprocessing, we index unique database values using LSH. Then, during the entity retrieval step, we query this index to quickly find the top similar values for a keyword.


\vspace{-0.7em}
\hypertarget{subsubsec:vector_database}{\paragraph{Vector Database for Descriptions.}}

As explained in \hyperlink{subsubsec:context_retrieval}{context retrieval}, extracting the most semantically relevant pieces of information from database catalogs is crucial for writing a SQL query. These documents can be extensive, with hundreds of pages explaining the entities and their relationships within the database, necessitating an efficient retrieval method.
To perform a high-efficiency semantic similarity search, we preprocess the database catalogs into a vector database. During the context retrieval step, we query this vector database to find the most relevant pieces of information for the question at hand.
For a more detailed description of our pipeline and the preprocessing phase, we encourage the reader to read Appendix \ref{sec:implementation_details}.

\section{Experiments}
\label{experiments}

\subsection{Datasets and Metrics}

The Spider dataset \citep{yu2018spider} includes 200 database schemas, with 160 schemas available for training and development, and 40 schemas reserved for testing. Notably, the databases used in the training, development, and test sets are distinct and do not overlap.

The recently introduced BIRD dataset \citep{li2024can} features 12,751 unique question-SQL pairs, covering 95 large databases with a combined size of 33.4 GB. This dataset spans 37 professional fields, including sectors such as blockchain, hockey, healthcare, and education. BIRD enhances SQL query generation by incorporating external knowledge and providing a detailed database catalog that includes column and database descriptions, thereby clarifying potential ambiguities. The SQL queries in BIRD are generally more complex than those found in the Spider dataset.
\vspace{-0.7em}
\hypertarget{SDS}{\paragraph{Subsampled Development Set (SDS).}}
To facilitate ablation studies, reduce costs, and maintain the distribution of the BIRD development set, we subsampled 10\% of each database in the development set, resulting in the Subsampled Development Set which we call SDS. This SDS consists of 147 samples: 81 simple, 54 moderate, and 12 challenging questions. For reproducibility, we included the SDS in our GitHub repository\footnote{\url{https://github.com/ShayanTalaei/CHESS}}.

\vspace{-0.7em}
\paragraph{Metrics.}
The Spider dataset evaluates SQL queries using two metrics: exact-set-match accuracy (EM) and execution accuracy (EX). Exact-set-match accuracy assesses each clause independently, requiring a perfect match with its corresponding clause in the reference SQL query. A SQL query is deemed correct only if it aligns completely with the reference across all components. However, this metric does not account for data values and has a high false negative rate due to multiple valid SQL formulations for a single question.

Execution accuracy (EX) evaluates the accuracy of the SQL output by comparing the results of the predicted query with those of the reference query when executed on specific database instances. This metric provides a more nuanced understanding of performance by accounting for variations in valid SQL queries that may arise from the same question. 

Additionally, the BIRD benchmark introduces the Valid Efficiency Score (VES), which evaluates SQL query performance by considering both accuracy and execution speed. We achieve a similar ranking compared to other models on this metric as well but due to its high variance and dependence on the computational environment we exclude it from the current analysis. 

\subsection{BIRD Results}
Since the test set of the BIRD benchmark is not available, we conducted our ablations and performance evaluations on the development set. We assessed our proposed method using both 1)  proprietary and 2)  open-source models. In the first scenario, we utilized our fine-tuned DeepSeek Coder model for candidate generation, GPT-3.5-turbo for column filtering, and GPT-4-turbo for the remaining LLM calls. We refer to this as our \hypertarget{default_engine_setup}{default engine set up}. In the second scenario, we used our fine-tuned DeepSeek Coder model for candidate generation, with all other LLM calls handled by Llama-3-70B.

As reported in Table \ref{table: bird_dev}, our approach using proprietary models achieved state-of-the-art execution accuracy on both the development and test sets of BIRD. Our method with open-source LLMs attained the highest performance among all open-source methods. At the time of the submission of this paper, the highest-performing method on the BIRD leaderboard is ExSL + granite-20b-code with an accuracy of 67.86\% on the test set. Our approach ranks second with an accuracy of 66.69\%.

\begin{table}[ht]
  \caption{Performance of our proposed method on the BIRD development set and Spider test set, comparing to all published methods.}
  \label{table:combined}
  \centering
  \begin{tabular}{l@{\hspace{1.5cm}}r}
    \begin{subtable}[t]{0.45\textwidth}
      \centering
      \caption{BIRD results}
      \label{table: bird_dev}
      \begin{tabular}[t]{lcc}
        \toprule
        Method & test EX & dev EX   \\
        \midrule
        CHESS + proprietary (ours) &  \textbf{66.69} & \textbf{65.00} \\
        MCS-SQL + GPT-4 & 65.45 & 63.36  \\ 
        \citep{lee2024mcs} & & \\
        CHESS + Open LLMs (ours) & -- & \textbf{61.5} \\
        SFT CodeS-15B & 60.37  & 58.47  \\
        \citep{li2024codes} & & \\
        DTS-SQL + DeepSeek 7B & 60.31 & 55.8  \\
        \citep{pourreza2024dts} & & \\
        MAC-SQL + GPT-4 & 57.56 & 59.59 \\
        \citep{wang2023mac} & & \\
        \bottomrule
      \end{tabular}
    \end{subtable}
    &
    \begin{subtable}[t]{0.45\textwidth}
      \centering
      \caption{Spider test set}
      \label{table: spider}
      \begin{tabular}{lccc}
        \toprule
        Method   & EX  \\
        \midrule
        MCS-SQL+GPT-4 & 89.6 \\
        CHESS (ours) & \textbf{87.2}  \\
        DAIL-SQL + GPT-4 & 86.6   \\
        \citep{gao2023text} & \\
        DIN-SQL + GPT-4 & 85.3 \\
        \citep{pourreza2024din} & \\
        C3 + ChatGPT & 82.3 \\
        \citep{dong2023c3} & \\
        RESDSQL-3B & 79.9\\
        \citep{li2023resdsql} & \\
        \bottomrule
      \end{tabular}
    \end{subtable}
  \end{tabular}
\end{table}

\subsection{Spider Results}


To evaluate the generalizability of our proposed method beyond the BIRD benchmark, we tested it on the Spider test set without specifically fine-tuning a new model for candidate generation or modifying the in-context learning samples. We followed our \hyperlink{default_engine_setup}{default engine setup}. The only adjustment we made to our pipeline was the removal of the context retrieval node since the Spider test set lacks column or table descriptions, which are integral to our method. As shown in Table \ref{table: spider}, our approach achieved an execution accuracy of 87.2\% on 2,147 samples from the test set, ranking it as the second-highest performing method among those published. This underscores the robustness of our method across different databases without any modifications. Notably, the best propriety (and undisclosed) method on the Spider test set leaderboard is Miniseek with an accuracy of 91.2\%.


\subsection{Ablation Studies}
\label{ablations}

\hypertarget{subsubsec:models_ablation}{\paragraph{Models Ablation.}} 
Thanks to our efficient retrieval process, which carefully controls the number of tokens passed to LLMs, we can utilize an open-source LLM with a small context window size, specifically Llama-3 with only 8K tokens \cite{meta-llama}. This contrasts with previous works that predominantly use GPT-4 as their base model \citep{pourreza2024din, lee2024mcs, wang2023mac}. In Table \ref{table:engines_ablations}, we present the results of our proposed pipeline using various LLMs from different families on a \hyperlink{SDS}{subsampled development set} dataset. The results indicate that our fine-tuned model for candidate generation significantly enhances performance. Notably, Llama-3's performance surpasses that of GPT-3.5-turbo but does not yet reach the performance levels of GPT-4 in our analysis.

\begin{table}[ht]
  \caption{This table shows the execution accuracy (EX) of different engine setups on the subsampled development set. Each engine setup is represented as a triplet (column filtering, candidate query generation, table/column selection + revision).}
  \label{table:engines_ablations}
  \centering
  \resizebox{0.6\columnwidth}{!}{ 
    \begin{tabular}{lc}
      \toprule
      Engine setups  & EX   \\
      \midrule
      (GPT-3.5-turbo, Fine-tuned DeepSeek, GPT-4-turbo) & 64.62 \\
      (GPT-3.5-turbo, GPT-4-turbo, GPT-4-turbo)         & 55.78 \\
      (GPT-3.5-turbo, GPT-3.5-turbo, GPT-3.5-turbo)     & 49.65 \\
      (Llama-3-70B, Llama-3-70B, Llama-3-70B)           & 54.42 \\
      (Llama-3-70B, Fine-tuned DeepSeek, Llama-3-70B)   & 59.86 \\
      \bottomrule
    \end{tabular}
  }
\end{table}

\begin{table}[ht]
\caption{The execution accuracy (EX) of the pipeline by removing each component on the (subsampled) dev set.}
\label{table:pipeline_ablations}
\centering
\resizebox{0.5\columnwidth}{!}{
    \begin{tabular}{lcc}
    \toprule
    Pipeline Setup & EX & $\Delta$EX \\
    \midrule
    Full pipeline & 64.62 & -- \\
    w/o entity \& context retrieval & 59.86 & -4.76 \\
    w/o individual column filtering & 61.90 & -2.72 \\
    w/o table selection & 58.50 & -6.12 \\
    w/o final column selection & 59.18 & -5.44 \\
    w/o revision & 57.82 & -6.80 \\
    with 1-time revision & 61.22 & -3.40 \\
    \bottomrule
    \end{tabular}
}
\end{table}

\hypertarget{subsubsec:modules_ablation}{\paragraph{Modules Ablation.}} Table \ref{table:pipeline_ablations} presents the execution accuracy (EX), where different modules or components are omitted. In the configuration without entity and context retrieval, we retrieved a random example and included column descriptions for all columns.  This approach highlights the significant impact of our selective retrieval, which outperforms naive context augmentation by 4.76\% in execution accuracy. Additionally, we evaluated the effect of removing each submodule within the schema selection module, revealing that table selection is the most critical, contributing a 6.12\% increase in performance. The table also illustrates the significant influence of the revision node, with a 6.80\% improvement. Increasing the number of revision samples for self-consistency led to higher performance gains, aligning with findings from \citep{lee2024mcs}.

\subsection{Performance Evaluation Across Queries with Varying  Complexity}
\label{complexity_analysis}

The BIRD benchmark categorizes questions and SQL pairs based on the number and type of SQL keywords used into three classes: easy, moderate, and challenging. In this section, we evaluate  the performance of our method  1) using our fine-tuned model for candidate generation, and 2) using GPT-4 without our fine-tuned model. We compare the results to the original GPT-4 baseline as in the BIRD paper, where the question, evidence, and the complete schema with all tables and columns are presented to GPT-4 along with chain-of-thought reasoning prompts.

The analysis is conducted on the \hyperlink{SDS}{SDS} dataset, and the results are detailed in Table \ref{table:complexity_analysis}. Our proposed method in both settings significantly improved performance across all classes. This analysis further demonstrates that providing an LLM with all available information can confuse the model and selective retrieval is crucial for achieving higher performance.

\begin{table}[h]
    \caption{Comparing the performance of our proposed method in two different settings with naively passing all information to GPT-4 across different difficulty levels on the subsampled development set.}
    \centering
    \resizebox{0.7\columnwidth}{!}{
        \begin{tabular}{>{\raggedright\arraybackslash}m{4cm}cccc}
            \toprule
            & Easy & Moderate & Challenging & Overall \\
            \midrule
            CHESS (with fine-tuning) & 65.43 & 64.81 & 58.33 & 64.62 \\
            CHESS (w/o fine-tuning) & 60.49 & 50.00 & 50.00 & 55.78 \\
            GPT-4-turbo (baseline) & 54.32 & 35.18 & 41.66 & 46.25 \\
            \bottomrule
        \end{tabular}
    }
    \label{table:complexity_analysis}
\end{table}

\subsection{Evaluation of the Schema Selection}

Aside from the ablation studies to measure the effects of context retrieval, schema selection, and revision methods, we also measure their effect on the precision and recall of the tables and columns provided to the model for generating the final SQL. Precision and recall are calculated using the columns and tables that are used in the correct SQL queries as ground truth. As shown in Table \ref{table:recall_precision}, each step of the pipeline increases the precision of the selected tables and columns while only slightly decreasing the recall. 

By achieving high recall and precision, the model has the most relevant information to generate the correct SQL in a small context window. An example of the reduction in the number of tables and columns of our pipeline is shown in Figure \ref{fig:funnel_graph} where the final two tables are selected correctly and the final selected five columns include the two correct columns used by the correct SQL. 

\begin{table}[h]
    \caption{Recall and Precision for Individual Column Filtering, Table Selection, and Column Selection compared to the tables and columns used in the correct SQL.}
    \centering
    \resizebox{0.7\columnwidth}{!}{ 
        \begin{tabular}{>{\raggedright\arraybackslash}m{4cm}cccc}
            \toprule
            & \multicolumn{2}{c}{Table} & \multicolumn{2}{c}{Column} \\
            \cmidrule(lr){2-3} \cmidrule(lr){4-5}
            & Recall & Precision & Recall & Precision \\
            \midrule
            No Filtering and Selection & 1.0 & 0.33 & 1.0 & 0.11 \\
            Individual Column Filtering & 1.0 & 0.33 & 0.98 & 0.21 \\
            Table Selection & 0.97 & 0.89 & 0.96 & 0.45 \\
            Final Column Selection & 0.96 & 0.90 & 0.94 & 0.71 \\
            \bottomrule
        \end{tabular}
    }
    \label{table:recall_precision}
\end{table}

\section{Discussion and Limitations}
\label{sec:discussion_and_limitations}

In this paper, we propose a new LLM-powered pipeline, called CHESS, for effective text-to-SQL synthesis. CHESS consists of novel and efficient retrieval and schema pruning methodologies as well as scalable finetuning techniques. Our pipeline achieved state-of-the-art performance among all known methodologies on the challenging BIRD benchmark. Furthermore, we also developed an entirely open-source version of CHESS, that for the first time, surpassed 60\% execution accuracy on BIRD, narrowing the performance gap between closed-source and open-source LLMs, while ensuring the privacy of the data.

While our approach increases the capability of text-to-SQL methodologies, the ultimate goal is to bring full automation to the database querying process. For the challenging BIRD dataset, humans still achieve a larger performance on query synthesis and future work should aim to further close this gap. We have quantified the efficacy of each of the components of our pipeline, which presents a guideline for strategies for further enhancements. Each of the information retrieval, schema pruning, and synthesis could benefit from further improvements, although as described in the paper, devising higher precision schema selection methodologies would be a high-impact area for future research that as it stands, might have an outsize impact on end-to-end accuracy.


%

%




\newpage

\bibliographystyle{plainnat}
\bibliography{custom.bib}

 \newpage
\appendix
\input{appendix.tex}

\label{appendix}

\end{document}

%% file: appendix.tex
\onecolumn

\section{Implementation Details}
\label{sec:implementation_details}
\input{implementation_details.tex}

\section{Finetuning the Model for Generate\_Candidate\_Query}
\label{sec:finetuning_candidate_generator}
\input{finetuning.tex}

\newpage
\onecolumn
\section{Prompt Templates}
\label{sec:prompt_templates}
\input{prompt_templates.tex}

\clearpage



\section{Extended Experiments}
\label{sec:extended_experiments}
\input{extended_experiments.tex}

\clearpage

\section{Schema Selection Example}
\label{sec:schema_selection_example}
\input{ERD_example.tex}

\clearpage

\section{Error Analysis}
\label{sec:error_analysis}
\input{error_analysis.tex}

\clearpage

%% file: implementation_details.tex
\subsection{Locality Sensitive Hashing Indexing of Database Values}
Our goal in the \emph{retrieve\_entity} tool is to retrieve database values that most closely match a set of keywords derived from the question. It is important to recognize that keywords from the question may not exactly correspond to the database values due to potential typos, variations in expression, or the common scenario where users are unaware of the precise format used to store data in the database. This reality demands a retrieval strategy that is both robust and adaptable, capable of accommodating such discrepancies. Relying solely on exact match retrieval, as suggested in prior studies \citep{li2024codes}, may not be sufficiently effective.

To address this, we employ string similarity measures, such as edit distance and semantic embedding, to retrieve the values most similar to the keywords. However, computing the edit distance and embedding similarity for every keyword against all values in the database is computationally expensive and time-consuming. To balance efficiency and accuracy, we utilize a hierarchical retrieval method.

Locality Sensitive Hashing (LSH) is an efficient technique for approximate nearest neighbor searches, which allows us to retrieve the most similar values to a keyword in the database. In the pre-processing stage, we index unique values in the database using LSH. Then, in the \emph{retrieve\_entity} tool of the IR agent, we query this index to rapidly find the top similar values to a keyword. Our approach involves using LSH queries to retrieve the top 10 similar values, after which we compute the edit distance and semantic similarity between the keyword and these values to further refine the results.

To simultaneously utilize edit distance and embedding similarity, we first identify the top 10 values closest to each keyword based on cosine similarity between their embedding vectors (obtained using OpenAI text-embedding-3-small \cite{openai_embeddings}) and the keyword's embedding vector. We then filter out values that fall below a specific threshold. Finally, for each keyword and column, we retain only the value that has the smallest edit distance.

We observed a significant reduction in time complexity, from 5 minutes to 5 seconds, using this method compared to a naive approach of computing the edit distance for all unique values in the database on the fly. While computing edit distance is proportional to the size of the database, significantly increasing the time complexity for processing a single question, using LSH allows us to index values in the pre-processing step and, during entity retrieval, rapidly query the index to find the most similar values to a keyword in a much more time-efficient manner.

\subsection{Vector database}

Each database schema in the BIRD benchmark \cite{li2024can} includes detailed descriptions for columns, specifying the contents of each column and the values for categorical columns. Providing these descriptions to the model is essential for guiding the SQL query generation process. However, incorporating all descriptions in the prompt for the weaker open-source models can overwhelm the model, potentially leading to the generation of incorrect SQL queries, as observed in section \ref{complexity_analysis}. It is important to note that the database catalog in the BIRD benchmark provides a relatively limited view of database metadata. In contrast, real-world production-level databases often contain more diverse information, including value ranges, constraints, and usage instructions for each table. Our proposed method can effectively utilize this extensive metadata to enhance performance.

To evaluate the relevance of descriptions to a given question, we employ embedding similarity \citep{openai_embeddings}, which quantifies the semantic similarity between the question and each description. To enhance the efficiency of the retrieval process, we pre-process the descriptions and created the embedding vectors for each of them and stored in a vector database, utilizing ChromaDB in our implementation. For the \emph{retrieve\_context} tool of the IR agent, we query this vector database to identify descriptions that are most semantically aligned with the question. This targeted approach ensures that only the most pertinent information is provided to the model, thereby improving the accuracy of the generated SQL queries.

\subsection{Local Column filtering}

In the \emph{filter\_column} tool, decisions to retain a column for subsequent steps are made independently, without considering relative information. The data provided to the LLM for column filtering includes: 1) the table name, 2) the column name, 3) the data type, 4) descriptions, if retrieved by the IR agent, and 5) database values, if retrieved by \emph{retrieve\_entity} tool. To enhance the model performance and make the task definition clear to the model, we used few-shot samples for this tool.  

Some key columns for SQL generation, which we call linking columns, such as those with foreign and primary key constraints, are crucial for writing SQL queries. For instance, questions about counting entities often requires primary keys, and joining tables necessitates foreign key columns. However, in the \emph{filter\_column}, \emph{select\_table} \emph{select\_column} tools, some of these essential columns may be initially rejected because they do not semantically relate to the given question. Despite this, these columns are indispensable for SQL generation. Therefore, in all of our sub-modules, we consistently retain foreign key and primary key columns, irrespective of the outputs from column selection and filtering processes.

\subsection{Revise Tool}

Revising generated candidate SQL queries is a critical aspect for the Candidate Generator Agent. In addition to the database schema, the question, and the candidate SQL query, we also provide the execution result of the SQL query. This gives the LLM an opportunity to view the retrieved data and revise the SQL query accordingly. This process mirrors human behavior when writing complex SQL queries; typically, we start with a draft query and refine it based on the results of its execution. Furthermore, this method allows the LLM to make necessary adjustments to the SQL query in instances of execution syntax errors.

\begin{figure*}[h]
    \centering
    \includegraphics[width=0.8\textwidth]{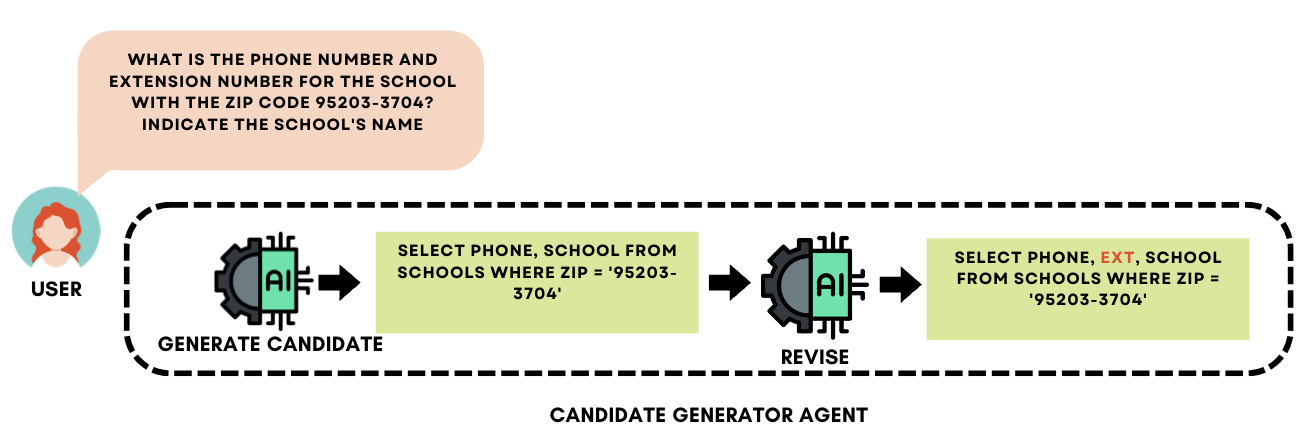}
    \caption{An example of the revise tool to fix missing columns in a candidate query.}
    \label{fig:revision_example}
\end{figure*}

In this step, we also incorporate instructions derived from our error analysis \ref{sec:error_analysis} to guide the model towards generating correct SQL queries. For instance, as shown in \ref{fig:revision_example}, we guided the model to ensure that all requested columns are included in the SQL query. In this specific example, the \emph{revise} tool identified a missing column and successfully added it to the query.

\newpage

%% file: finetuning.tex
\label{finetuning_experiments}

\subsection{Fine-tuning Dataset and Model}

To enhance the generation of better candidate SQL queries for the Candidate Generator Agent, we fine-tuned the DeepSeek Coder 34B \cite{guo2024deepseek} on the training set of the BIRD benchmark, which comprises approximately 9,500 samples. In constructing the fine-tuning dataset, rather than solely using correct tables and columns like the work proposed in \cite{pourreza2024dts}, we developed a heuristic to address the error propagation issue. Recognizing that previous steps in the pipeline may not always pinpoint the most efficient schema with perfect accuracy, we intentionally introduced some noise into our dataset creation to train the model. 
Specifically, we included columns and tables that were incorrect but shared similar naming conventions and semantic attributes with the correct schema. We also utilized our keyword selection module to extract keywords from the questions, search for these keywords in the database, and incorporate them into the prompt.

\subsection{Hyperparameters}

We fine-tuned the deepseek model for the \emph{generate\_candidate\_query} tool using 4-bit quantization of the base model and LORA adapters \cite{hu2021lora}, a technique formally referred to as QLORA \cite{dettmers2024qlora}. We configured the LORA rank parameter to 128 and set the LORA alpha parameter to 256. The fine-tuning process was conducted over two epochs on the constructed dataset, utilizing a batch size of 32 and a learning rate of 1e-4, along with a cosine scheduler, all on a single H100 GPU for 4 hours.

%% file: prompt_templates.tex
In this section we provide the exact prompts that have been used for the tools in our multi-agent setup. For easier parsing of the LLM's output, we instruct the model to generate its output in a structured format.

\begin{figure}[h]
\centering
\begin{tikzpicture}
  \node[draw, rectangle, inner sep=2mm, fill=mylightblue] (rect) {
    \begin{minipage}{\linewidth-4mm}
      \texttt{Objective: Analyze the given question and hint to identify and extract keywords, keyphrases, and named entities. These elements are crucial for understanding the core components of the inquiry and the guidance provided. This process involves recognizing and isolating significant terms and phrases that could be instrumental in formulating searches or queries related to the posed question.\\
      \\
      Instructions:\\
      1. Read the Question Carefully: Understand the primary focus and specific details of the question. Look for any named entities (such as organizations, locations, etc.), technical terms, and other phrases that encapsulate important aspects of the inquiry.\\
      2. Analyze the Hint: The hint is designed to direct attention toward certain elements relevant to answering the question. Extract any keywords, phrases, or named entities that could provide further clarity or direction in formulating an answer.\\
      3. List Keyphrases and Entities: Combine your findings from both the question and the hint into a single Python list. This list should contain:\\
      - Keywords: Single words that capture essential aspects of the question or hint.\\
      - Keyphrases: Short phrases or named entities that represent specific concepts, locations, organizations, or other significant details.\\
      Ensure to maintain the original phrasing or terminology used in the question and hint.\\
      \\
      \{FEWSHOT\_EXAMPLES\}\\
      \\
      Task:\\
      Given the following question and hint, identify and list all relevant keywords, keyphrases, and named entities.\\
      \\
      Question: \{QUESTION\}\\
      \\
      Hint: \{HINT\}\\
      \\
      Please provide your findings as a Python list, capturing the essence of both the question and hint through the identified terms and phrases.\\
      Only output the Python list, no explanations needed.
      }
    \end{minipage}
  };
\end{tikzpicture}
\caption{Template for the \emph{extract\_keyword} tool}
\label{fig:keywordextraction}
\end{figure}
  
\begin{figure}[h]
\centering
\begin{tikzpicture}
  \node[draw, rectangle, inner sep=2mm, fill=mylightblue] (rect) {
    \begin{minipage}{\linewidth-4mm}
      \texttt{You are an expert and very smart data analyst.\\
      Your task is to analyze the provided database schema, comprehend the posed question, and leverage the hint to identify which tables are needed to generate a SQL query for answering the question.\\
      \\ Database Schema Overview:\\
      \{DATABASE\_SCHEMA\}\\
      \\
      This schema provides a detailed definition of the database's structure, including tables, their columns, primary keys, foreign keys, and any relevant details about relationships or constraints.\\
      For key phrases mentioned in the question, we have provided the most similar values within the columns denoted by "-- examples" in front of the corresponding column names. This is a critical hint to identify the tables that will be used in the SQL query.\\
      \\Question:\\
      \{QUESTION\}\\
      \\
      Hint:\\
      \{HINT\}\\
      \\
      The hint aims to direct your focus towards the specific elements of the database schema that are crucial for answering the question effectively.\\
      \\
      Task:\\
      Based on the database schema, question, and hint provided, your task is to determine the tables that should be used in the SQL query formulation. \\
      For each of the selected tables, explain why exactly it is necessary for answering the question. Your explanation should be logical and concise, demonstrating a clear understanding of the database schema, the question, and the hint.\\
      \\
      Please respond with a JSON object structured as follows:\\
      \\
      \{\\
      \hspace*{1em}"chain\_of\_thought\_reasoning": "Explanation of the logical analysis that led to the selection of the tables.",\\
      \hspace*{1em}"table\_names": ["Table1", "Table2", "Table3", ...]\\
      \}\\
      \\
      Note that you should choose all and only the tables that are necessary to write a SQL query that answers the question effectively.\\
      Only output a json as your response.
      }
    \end{minipage}
  };
\end{tikzpicture}
\caption{Template for the \emph{select\_tables} tool.}
\label{fig:table_selection}
\end{figure}

\begin{figure}[H] 
    \centering
    \begin{tikzpicture}
      \node[draw, rectangle, inner sep=2mm, fill=mylightblue] (rect) {
        \begin{minipage}{\linewidth-2mm}
          \texttt{You are an expert and very smart data analyst.\\
          Your task is to examine the provided database schema, understand the posed question, and use the hint to pinpoint the specific columns within tables that are essential for crafting a SQL query to answer the question.\\
          \\Database Schema Overview:\\
          \{DATABASE\_SCHEMA\}\\
          \\
          This schema offers an in-depth description of the database's architecture, detailing tables, columns, primary keys, foreign keys, and any pertinent information regarding relationships or constraints. Special attention should be given to the examples listed beside each column, as they directly hint at which columns are relevant to our query.\\
          For key phrases mentioned in the question, we have provided the most similar values within the columns denoted by "-- examples" in front of the corresponding column names. This is a critical hint to identify the columns that will be used in the SQL query.\\
          \\Question:\\
          \{QUESTION\}\\
          \\
          Hint:\\
          \{HINT\}\\
          \\
          The hint aims to direct your focus towards the specific elements of the database schema that are crucial for answering the question effectively.\\
          \\
          Task:\\
          Based on the database schema, question, and hint provided, your task is to identify all and only the columns that are essential for crafting a SQL query to answer the question.\\
          For each of the selected columns, explain why exactly it is necessary for answering the question. Your reasoning should be concise and clear, demonstrating a logical connection between the columns and the question asked.\\
          \\
          Tip: If you are choosing a column for filtering a value within that column, make sure that column has the value as an example.\\
          \\
          Please respond with a JSON object structured as follows:\\
          \\
          \{\\
          \hspace*{1em}"chain\_of\_thought\_reasoning": "Your reasoning for selecting the columns, be concise and clear.",\\
          \hspace*{1em}"table\_name1": ["column1", "column2", ...],\\
          \hspace*{1em}"table\_name2": ["column1", "column2", ...],\\
          \hspace*{1em}...\\
          \}\\
          \\
          Make sure your response includes the table names as keys, each associated with a list of column names that are necessary for writing a SQL query to answer the question.\\
          For each aspect of the question, provide a clear and concise explanation of your reasoning behind selecting the columns.\\
          Only output a json as your response.
          }
        \end{minipage}
      };
    \end{tikzpicture}
    \caption{Template for the \emph{select\_columns} tool.}
    \label{fig:column_selection}
\end{figure}

  \begin{figure}[h]
    \centering
    \begin{tikzpicture}
      \node[draw, rectangle, inner sep=2mm, fill=mylightblue] (rect) {
        \begin{minipage}{\linewidth-4mm}
          \texttt{You are a detail-oriented data scientist tasked with evaluating the relevance of database column information for answering specific SQL query question based on provided hint.\\
          Your goal is to assess whether the given column details are pertinent to constructing an SQL query to address the question informed by the hint. Label the column information as "relevant" if it aids in query formulation, or "irrelevant" if it does not.\\
          \\
          Procedure:\\
          1. Carefully examine the provided column details.\\
          2. Understand the question about the database and its associated hint.\\
          3. Decide if the column details are necessary for the SQL query based on your analysis.\\
          \\
          Here is an example of how to determine if the column information is relevant or irrelevant to the question and the hint:\\
          \\
          \{FEWSHOT\_EXAMPLES\}\\
          \\
          Now, it's your turn to determine whether the provided column information can help formulate a SQL query to answer the given question, based on the provided hint.\\
          \\
          The following guidelines are VERY IMPORTANT to follow. Make sure to check each of them carefully before making your decision:\\
          1. You're given only one column's information, which alone isn't enough to answer the full query. Concentrate solely on this provided data and assess its relevance to the question and hint without considering any missing information.\\
          2. Read the column information carefully and understand the description of it, then see if the question or the hint is asking or referring to the same information. If yes then the column information is relevant, otherwise it is irrelevant.\\
          ...\\
          \\
          Column information:\\
          \{COLUMN\_PROFILE\}\\
          \\
          Question:\\
          \{QUESTION\}\\
          \\
          HINT:\\
          \{HINT\}\\
          \\
          Take a deep breath and provide your answer in the following json format:\\
          \\
          \{
            \hspace*{1em}"chain\_of\_thought\_reasoning": "One line explanation of why or why not the column information is relevant to the question and the hint.",\\
            \hspace*{1em}"is\_column\_information\_relevant": "Yes" or "No"\\
          \}\\
          \\
          Only output a json as your response.
          }
        \end{minipage}
      };
    \end{tikzpicture}
    \caption{Template for the \emph{filter\_column}.}
    \label{fig:column_filtering}
  \end{figure}
  
  \begin{figure}[h]
    \centering
    \begin{tikzpicture}
      \node[draw, rectangle, inner sep=2mm, fill=mylightblue] (rect) {
        \begin{minipage}{\linewidth-4mm}
          \texttt{You are a data science expert.\\
          Below, you are presented with a database schema and a question.\\
          Your task is to read the schema, understand the question, and generate a valid SQLite query to answer the question.\\
          Before generating the final SQL query think step by step on how to write the query.\\
          \\
          Database Schema:\\
          \{DATABASE\_SCHEMA\}\\
          \\
          This schema offers an in-depth description of the database's architecture, detailing tables, columns, primary keys, foreign keys, and any pertinent information regarding relationships or constraints. Special attention should be given to the examples listed beside each column, as they directly hint at which columns are relevant to our query.\\
          \\
          Database admin instructions:\\
          - Make sure you only output the information that is asked in the question. If the question asks for a specific column, make sure to only include that column in the SELECT clause, nothing more.\\
        - Predicted query should return all of the information asked in the question without any missing or extra information.\\
          \\
          Question:\\
          \{QUESTION\}\\
          \\
          Hint:\\
          \{HINT\}\\
          \\
          Please respond with a JSON object structured as follows:\\
          \\
          \{\\
          \hspace*{1em}"chain\_of\_thought\_reasoning": "Your thought process on how you arrived at the final SQL query.",\\
          \hspace*{1em}"SQL": "Your SQL query in a single string."\\
          \}\\
          \\
          Priority should be given to columns that have been explicitly matched with examples relevant to the question's context.\\
          \\
          Take a deep breath and think step by step to find the correct SQLite SQL query.
          }
        \end{minipage}
      };
    \end{tikzpicture}
    \caption{Template for the \emph{generate\_candidate\_query} tool}
    \label{fig:candidate_generation}
  \end{figure}

  \begin{figure}[h]
    \centering
    \begin{tikzpicture}
      \node[draw, rectangle, inner sep=2mm, fill=mylightblue] (rect) {
        \begin{minipage}{\linewidth-4mm}
          \texttt{Objective: Your objective is to make sure a query follows the database admin instructions and use the correct conditions.\\
          \\
          Database Schema:\\
          \{DATABASE\_SCHEMA\}\\
          \\
          Database admin instructions:\\
          - Make sure you only output the information that is asked in the question. If the question asks for a specific column, make sure to only include that column in the SELECT clause, nothing more.\\
        - Predicted query should return all of the information asked in the question without any missing or extra information.\\
          \\
          \{MISSING\_ENTITIES\}\\
          \\
          Question:\\
          \{QUESTION\}\\
          \\
          Hint:\\
          \{EVIDENCE\}\\
          \\
          Predicted query:\\
          \{SQL\}\\
          \\
          Query result:\\
          \{QUERY\_RESULT\}\\
          \\
          Please respond with a JSON object structured as follows (if the sql query is correct, return the query as it is):\\
          \\
          \{\\
          \hspace*{1em}"chain\_of\_thought\_reasoning": "Your thought process on how you arrived at the solution. You don't need to explain the instructions that are satisfied.",\\
          \hspace*{1em}"revised\_SQL": "Your revised SQL query."\\
          \}\\
          \\
          Take a deep breath and think step by step to find the correct SQLite SQL query.}
        \end{minipage}
      };
    \end{tikzpicture}
    \caption{Template for \emph{revise} tool}
    \label{fig:revision}
  \end{figure}

  \begin{figure}[h]
    \centering
    \begin{tikzpicture}
      \node[draw, rectangle, inner sep=2mm, fill=mylightblue] (rect) {
        \begin{minipage}{\linewidth-4mm}
            \texttt{** Instructions: **\\
            \\
            Given the following question database schema, and candidate responses, generate a set of {UNIT\_TEST\_CAP} unit tests that would evaluate the correctness of SQL queries that would answer the question. \\
            Unit tests should be designed in a way that distinguish the candidate responses from each other.\\
            \\
            - The unit tests should cover various aspects of the question and ensure comprehensive evaluation. \\
            - Each unit test should be clearly stated and should include the expected outcome.\\
            - Each unit test should be designed in a way that it can distinguishes at lease two candidate responses from each other.\\
            - The unit test should be formatted like 'The answer SQL query should mention...', 'The answer SQL query should state...', 'The answer SQL query should use...', etc. followed by the expected outcome.\\
            - First think step by step how you can design the units tests to distinguish the candidate responses using the <Thinking> tags.\\
            - After the thinking process, provide the list of unit tests in the \textless Answer\textgreater tags.\\
            \\
            VERY IMPORTANT:\\
            All of the unit tests should consider the logic of the SQL query do not consider the formatting of the output or output values. \\
            \\
            You are provided with different clusters of the candidate responses. Each cluster contains similar responses based on their results.\\
            You MUST generate test cases that can distinguish between the candidate responses in each cluster and the test case should promote the candidate responses that you think are correct.\\
            \\
            Example of the output format:\\
            \textless Thinking\textgreater\ Your step-by-step reasoning here. \textless Thinking\textgreater\\
            \textless Answer\textgreater\\
            \texttt{['The answer SQL query should mention...', ...]}\\
            \textless Answer\textgreater
            \\
            *** Database Schema: ** \\
            \{DATABASE\_SCHEMA\}\\
            \\
            *** Candidate Clusters: **\\
            \{CANDIDATE\_QUERIES\}\\
            \\
            *** Question: **\\
            Question: \{QUESTION\} (Hint: \{HINT\})\\}
        \end{minipage}
      };
    \end{tikzpicture}
    \caption{Template for \emph{generate\_unit\_tests} tool}
    \label{fig:unit_test_generation}
  \end{figure}

   \begin{figure}[h]
    \centering
    \begin{tikzpicture}
      \node[draw, rectangle, inner sep=2mm, fill=mylightblue] (rect) {
        \begin{minipage}{\linewidth-4mm}
        \texttt{
            ** Instructions: ** \\
Given the following question, database schema, a candidate SQL query response, and unit tests, evaluate whether or not the response passes each unit test.\\
\\
- In your evaluation, you should consider how the responses align with the a given unit test.\\
- Provide reasoning before you return your evaluation inside the \textless Thinking\textgreater tags.\\
- At the end of your evaluation, you must finish with a list of verdicts corresponding to each candidate responses in \textless Answer\textgreater tags. \\
- You must include a verdict with one of these formatted options: 'Passed' or 'Failed' \\
- Here is an example of the output format: \\
\textless Thinking\textgreater Your step by step reasoning here. \\ \textless Thinking\textgreater\\
\textless Answer\textgreater  \\
Candidate Response \#1: Passed \\
Candidate Response \#2: Failed \\
Candidate Response \#3: Passed \\
.... \\
\textless Answer\textgreater \\
- Each verdict should be on a new line and correspond to the candidate response in the same order as they are provided. \\
- Here is the question, database schema, candidate responses, and the unit test to evaluate the responses: \\
\\
*** Database Schema: ** \\
\{DATABASE\_SCHEMA\}\\
\\
*** Candidate Clusters: **\\
\{CANDIDATE\_QUERIES\}\\
\\
*** Question: **\\
Question: \{QUESTION\} (Hint: \{HINT\})\\
\\
*** Unit Test: **\\
\{UNIT\_TEST\}\\
}
        \end{minipage}
      };
    \end{tikzpicture}
    \caption{Template for \emph{evaluate} tool}
    \label{fig:unit_test_evaluation}
  \end{figure}

\begin{figure}[h]
    \centering
    \includegraphics[width=0.7\textwidth]{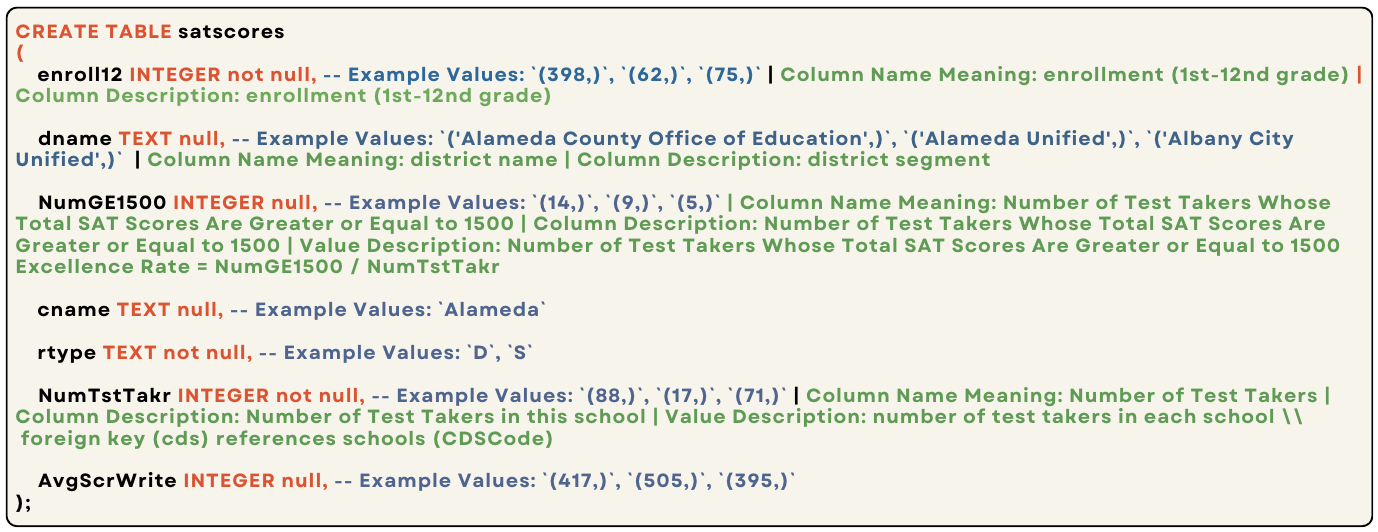}
    \caption{An example of how database schema, including the relevant values and descriptions, is formatted in the prompt.}
    \label{fig:db_schema_example}
\end{figure}

%% file: extended_experiments.tex
\subsection{Query Generation  with the Correct Context}

To maintain high performance while ensuring privacy, we fine-tuned an open-source model for the \emph{generate\_candidate\_query} tool of the Candidate Generator agent. This tuning incorporated the correct contextual information, including only relevant columns, tables, and their descriptions. As shown in \ref{tab: upper_bound}, using precise contextual information aligned with the gold SQL improved performance to 72.4\%. This result highlights the importance of retrieving efficient schema information, especially for open-source models, which struggle when handling excessively large input prompts.

\begin{table}[h]
  \caption{This table shows the maximum execution accuracy (EX) possible for our candidate SQL module generation by passing it the correct context for questions in the BIRD dataset.}
  \label{tab: upper_bound}
  \centering
  \begin{tabular}{lc}
    \toprule
    Engine  & EX   \\
    \midrule
    CHESS & 64.62 \\
    CHESS + correct context  & 72.4 \\
    \bottomrule
  \end{tabular}
\end{table}

\subsection{Models Ablation}
\label{subsubsec:models_ablation}
Thanks to our efficient retrieval process, which carefully controls the number of tokens passed to LLMs, we can utilize an open-source LLM with a small context window size, specifically Llama-3 with only 8K tokens \cite{meta-llama}. This contrasts with previous works that predominantly use GPT-4 as their base model \citep{pourreza2024din, lee2024mcs, wang2023mac}. In Table \ref{table:engines_ablations}, we present the results of our proposed framework using various LLMs from different families on a \hyperlink{SDS}{subsampled development set} dataset. The results indicate that our fine-tuned model for candidate generation significantly enhances performance. Notably, Llama-3's performance surpasses that of GPT-3.5-turbo but does not yet reach the performance levels of GPT-4 in our analysis.

\begin{table}[ht]
  \caption{This table shows the execution accuracy (EX) of different engine setups on the subsampled development set. Each engine setup is represented as a triplet (column filtering, candidate query generation, table/column selection + revision).}
  \label{table:engines_ablations}
  \centering
  \resizebox{0.5\columnwidth}{!}{ 
    \begin{tabular}{lc}
      \toprule
      Engine setups  & EX   \\
      \midrule
      (GPT-3.5-turbo, Fine-tuned DeepSeek, GPT-4-turbo) & 64.62 \\
      (GPT-3.5-turbo, GPT-4-turbo, GPT-4-turbo)         & 55.78 \\
      (GPT-3.5-turbo, GPT-3.5-turbo, GPT-3.5-turbo)     & 49.65 \\
      (Llama-3-70B, Llama-3-70B, Llama-3-70B)           & 54.42 \\
      (Llama-3-70B, Fine-tuned DeepSeek, Llama-3-70B)   & 59.86 \\
      \bottomrule
    \end{tabular}
  }
\end{table}

%% file: ERD_example.tex
\begin{figure}[h]
    \centering
    \includegraphics[width=0.4\textwidth]{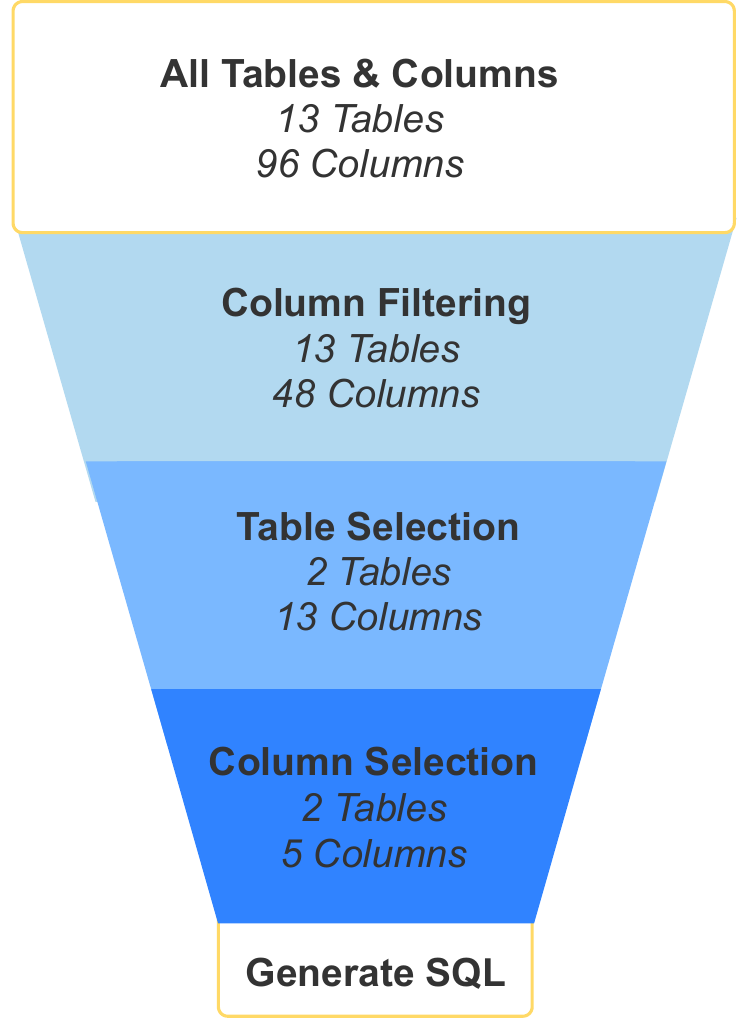}
    \caption{Funnel graph illustrating the progressive narrowing down of database schema with SS agent, leading to the final schema used for SQL generation.}
    \label{fig:funnel_graph}
\end{figure}

 In this part, we use an example to showcase how we narrow down the initial schema throughout the \emph{filter\_column}, the \emph{select\_tables}, and the \emph{select\_columns}.\\
The example is chosen from \emph{Formula 1} database from the BIRD benchmark, with a total of 13 tables and 96 columns. Here is the question and its evidence:
\begin{itemize}
    \item \textbf{Question: } What's the fastest lap time ever in a race for Lewis Hamilton?
    \item \textbf{Evidence: } fastest lap time ever refers to min(fastestLapTime)
\end{itemize}
Figure \ref{fig:funnel_graph} shows the number of tables and columns that are considered as the sub-selected schema after each tool. Starting from 13 tables and 96 columns, these numbers reduced to 36 columns in 13 tables after the \emph{filter\_column} tool. Subsequently, \emph{select\_tables} narrowed it down further to 2 tables and 7 columns. Finally, \emph{select\_columns} yielded the final schema with 2 tables and 5 columns, which is used for the SQL generation.

\begin{figure}[h]
        \centering
        \includegraphics[width=0.7\textwidth]{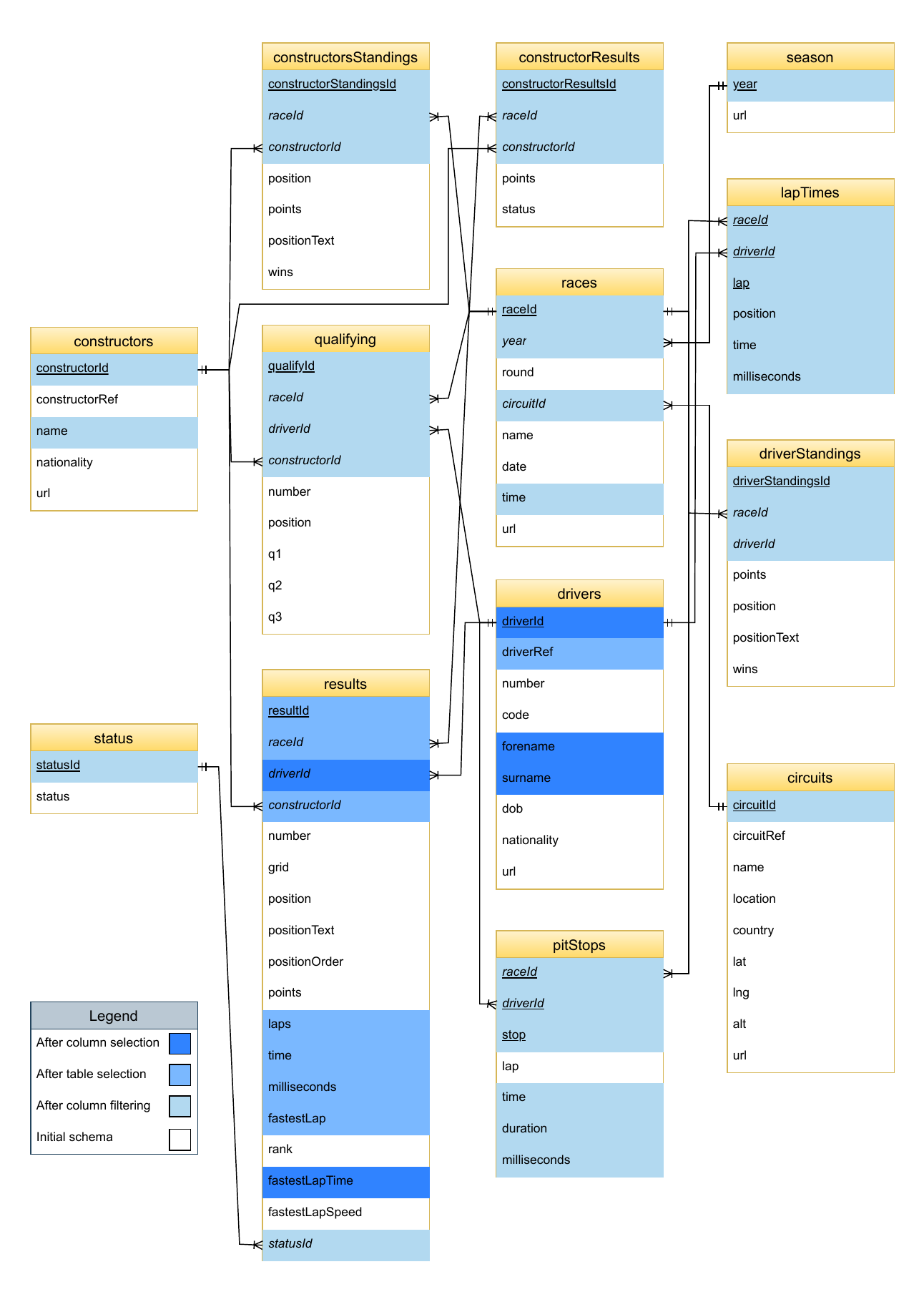}
        \caption{Schema selection example formula\_1\_926}
        \label{fig:ERD_example}
\end{figure}

To further illustrate the details of the schema selection process, we use the entity-relationship diagram (ERD) \ref{fig:ERD_example}. In this figure, each table is represented as a block with its columns listed below it. Primary keys are underlined and the foreign keys are in italics connecting the corresponding columns. As shown in the legend, the columns that remained in the selected schema after each tool are colored with a gradient of white to dark blue; white represents the columns that were present in the initial schema and got filtered after \emph{filter\_column} while dark blue shows the columns that are selected after \emph{select\_columns} to be passed to the Candidate Generator agent.\\
There are some points worth emphasizing in the plot. First, the decision to filter linking columns (Primary and foreign keys) is a task requiring a global view of the schema and cannot be done in the local view of column filtering, hence we do not filter these columns and include all of them in the result of column filtering step, explaining why all of the primary and foreign keys are present after this step. Second, some columns such as ``laps", ``time", and ``milliseconds" are all semantically related to the question because question asks for fastest lap time, which shows the \emph{filter\_column} tool successfully used the local information to find all relevant columns. However, all of these columns are not going to be used for crafting the SQL query, so we need to find the relevant columns with respect to their relative information, which is going to be done in the \emph{select\_tables} and \emph{select\_columns} tools. In the \emph{select\_tables} tool, as it can be observed from the figure, the ``lapTimes" table which has all information about time has been dropped by \emph{select\_tables} since there is a more relevant column, ``fastestLapTime", which can be used to answer the question. This was a concrete example, which shows how local and global view of the columns and tables can help to pinpoint the correct schema.

%% file: error_analysis.tex
To analyze our failure cases, we subsampled 147 questions from the development set of BIRD (\hyperlink{SDS}{SDS}) and processed these questions using our pipeline and a vanilla GPT-4 baseline. The vanilla GPT-4 baseline replicates the GPT-4 approach from BIRD, where the question, evidence, and the full schema with all tables and columns are provided to GPT-4 with chain-of-thought reasoning prompts. In this context, the evidence refers to the hint provided alongside some questions in the dataset.

\begin{figure*}[h] 
    \centering
    \begin{subfigure}[b]{0.45\textwidth}
        \centering
        \includegraphics[width=\linewidth]{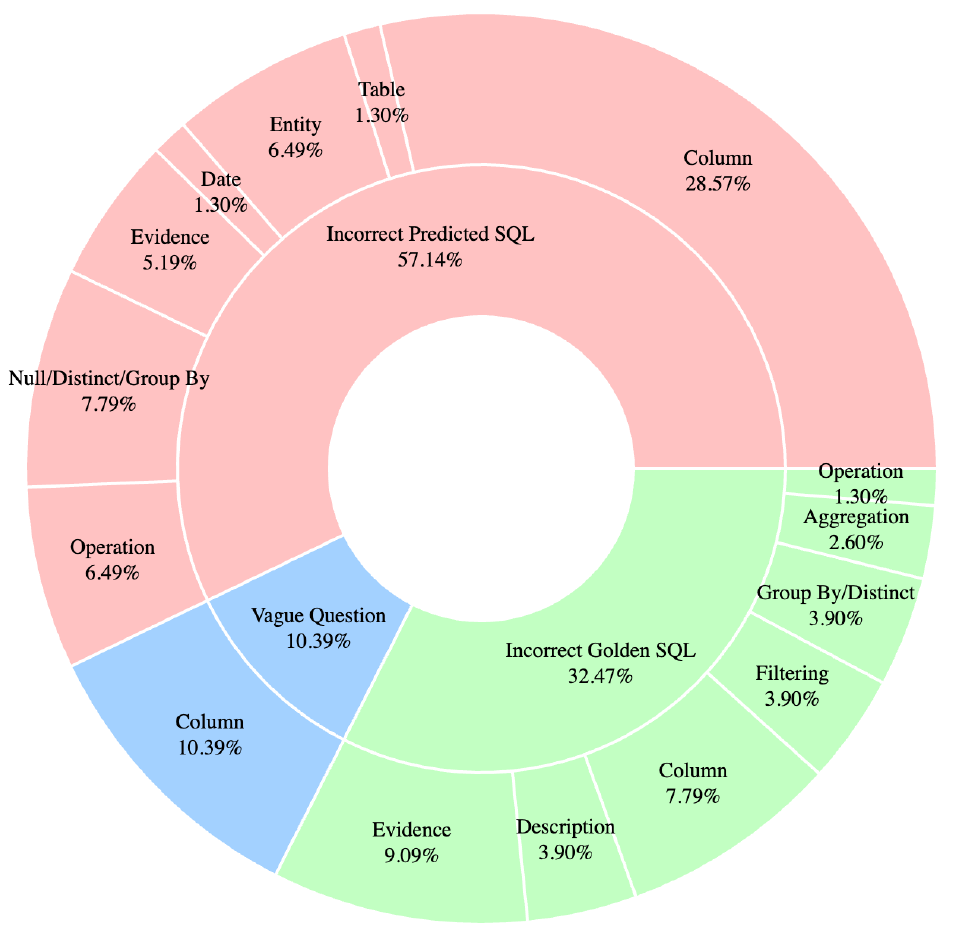}
        \caption{Vanilla GPT4}
        \label{fig:vanilla_gpt4_ea}
    \end{subfigure}
    \hfill 
    \begin{subfigure}[b]{0.45\textwidth}
        \centering
        \includegraphics[width=\linewidth]{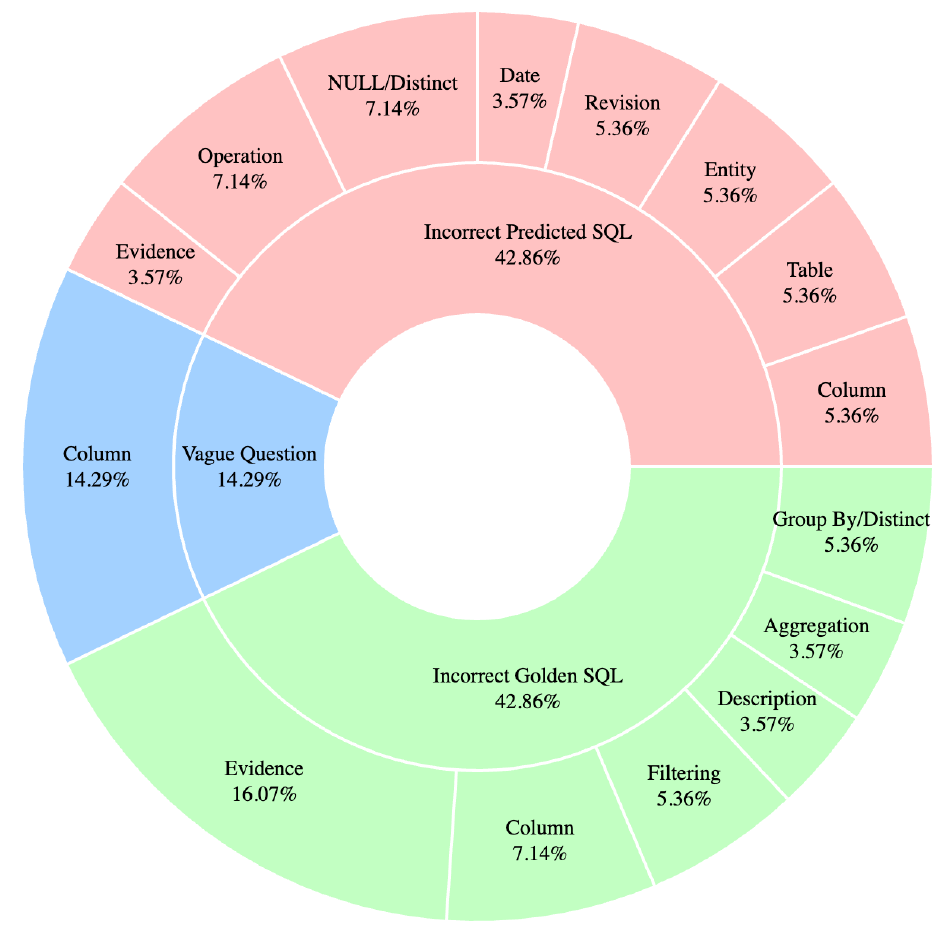}
        \caption{CHESS}
        \label{fig:chess_pipeline}
    \end{subfigure}
    \caption{Distribution of Errors on Sampled Dev Set.}
    \label{fig:side_by_side}
\end{figure*}

Figure~\ref{fig:side_by_side} shows the categories of errors and their percentages for our approach and the baseline. "Incorrectly Predicted SQL" refers to failures in our pipeline that lead to incorrect final SQL, while ambiguous questions and incorrect golden SQL indicate problems with the dataset, where the data formats expected from the questions were ambiguous, and the golden SQL provided is incorrect.

For the vanilla baseline, 57.1\% of the incorrectly predicted SQL results from incorrect schema linking issues, contributing 26.0\% of the errors due to wrong columns in SELECT or JOIN. In comparison, our approach has 42.9\% incorrect SQL, with only 5.4\% of the errors attributed to incorrect column linking. These figures show that our approach distributes error types more evenly, indicating improvements in all categories of potential errors.

\renewcommand{\arraystretch}{1.5}
\setlength{\extrarowheight}{1pt}

\newpage
\subsection{Incorrectly Predicted SQL}
\subsubsection{Column: }
Description: The columns used in the predicted SQL are either missing or incorrect.
\begin{longtable}{p{0.2\textwidth} p{0.8\textwidth}}
\hline
\hline 
\textbf{Question} & Which drivers born after 1975 have been ranked 2? Please give their forenames and surnames. \\
\hline
\hline 
\textbf{Evidence} & born after 1975 refers to year(dob) \textgreater 1975; \\
\hline
\textbf{Gold SQL} &
\texttt{SELECT T2.forename, T2.surname \newline
FROM results AS T1 \newline
INNER JOIN drivers AS T2 \newline
ON T1.driverId = T2.driverId \newline
WHERE STRFTIME('\%Y', T2.dob) > '1975' \newline
AND \textcolor{blue}{T1.rank = 2}} \\
\hline
\textbf{Predicted SQL} &
\texttt{SELECT T1.forename, T1.surname \newline
FROM drivers AS T1 \newline
INNER JOIN driverStandings AS T2 \newline
ON T1.driverId = T2.driverId \newline
WHERE \textcolor{red}{T2.position = 2} \newline
AND STRFTIME('\%Y', T1.dob) > '1975'} \\
\hline 
\textbf{Error Explanation} & "rank" is the correct column. \\
\hline
\hline
\hline 
\textbf{Question} & Find all cards illustrated by Stephen Daniel and describe the text of the ruling of these cards. State if these cards have missing or degraded properties and values. \\
\hline
\hline 
\textbf{Evidence} & cards have missing or degraded properties and value refers to hasContentWarning = 1; 'Stephen Daniele' is artist; \\
\hline
\textbf{Gold SQL} &
\texttt{SELECT \textcolor{blue}{T1.id}, T2.text, T1.hasContentWarning \newline
FROM cards AS T1 \newline
INNER JOIN rulings AS T2 \newline
ON T1.uuid = T2.uuid \newline
WHERE T1.artist = 'Stephen Daniele'} \\
\hline
\textbf{Predicted SQL} &
\texttt{SELECT \textcolor{red}{T1.uuid}, T2.text, T1.hasContentWarning \newline
FROM cards AS T1 \newline
INNER JOIN rulings AS T2 \newline
ON T1.uuid = T2.uuid \newline
WHERE T1.artist = 'Stephen Daniele'} \\
\hline 
\textbf{Error Explanation} & "id" is the correct column. \\
\hline
\end{longtable}

\subsubsection{Table: }
Description: The tables used in the predicted SQL are either missing or incorrect.
\begin{longtable}{p{0.2\textwidth} p{0.8\textwidth}}
\hline
\hline 
\textbf{Question} & Among the schools with the average score in Math over 560 in the SAT test, how many schools are directly charter-funded? \\
\hline
\hline 
\textbf{Gold SQL} &
\texttt{SELECT COUNT(T2.School Code) \newline
FROM satscores AS T1 \newline
INNER JOIN \textcolor{blue}{frpm} AS T2 \newline
ON T1.cds = T2.CDSCode \newline
WHERE T1.AvgScrMath > 560 \newline
AND \textcolor{blue}{T2.Charter Funding Type = 'Directly funded'}} \\
\hline
\textbf{Predicted SQL} &
\texttt{SELECT COUNT(T1.cds) \newline
FROM satscores AS T1 \newline
INNER JOIN \textcolor{red}{schools} AS T2 \newline
ON T1.cds = T2.CDSCode \newline
WHERE T1.AvgScrMath > 560 \newline
AND \textcolor{red}{T2.FundingType = 'Charter'}} \\
\hline 
\textbf{Error Explanation} & "frpm" is the correct table with the correct column. \\
\hline
\hline
\hline 
\textbf{Question} & Who placed the order with the id 32423? \\
\hline
\hline 
\textbf{Gold SQL} &
\texttt{SELECT T3.client\_id \newline
FROM order AS T1 \newline
INNER JOIN \textcolor{blue}{account} AS T2 \newline
ON T1.account\_id = T2.account\_id \newline
INNER JOIN \textcolor{blue}{client} AS T3 \newline
ON \textcolor{blue}{T2.district\_id = T3.district\_id} \newline
WHERE T1.order\_id = 32423} \\
\hline
\textbf{Predicted SQL} &
\texttt{SELECT T2.client\_id \newline
FROM \textcolor{red}{order} AS T1 \newline
INNER JOIN \textcolor{red}{disp} AS T2 \newline
ON T1.account\_id = T2.account\_id \newline
WHERE T1.order\_id = 32423} \\
\hline
\textbf{Error Explanation} & "account" and "client" are the correct tables with the correct columns. \\
\hline
\end{longtable}

\subsubsection{Entity:}
Description: The entity used for filtering is incorrect in the predicted SQL.
\begin{longtable}{p{0.2\textwidth} p{0.8\textwidth}}
\hline
\hline 
\textbf{Question} & For the customers who paid in the euro, what is their average total price of the transactions? \\
\hline
\hline 
\textbf{Gold SQL} &
\texttt{SELECT AVG(T1.Price) \newline
FROM transactions\_1k AS T1 \newline
INNER JOIN gasstations AS T2 \newline
ON T1.GasStationID = T2.GasStationID \newline
INNER JOIN customers AS T3 \newline
ON T1.CustomerID = T3.CustomerID \newline
WHERE T3.Currency = \textcolor{blue}{'EUR'}} \\
\hline
\textbf{Predicted SQL} &
\texttt{SELECT AVG(T1.Price) \newline
FROM transactions\_1k T1 \newline
JOIN customers T2 \newline
ON T1.CustomerID = T2.CustomerID \newline
WHERE T2.Currency = \textcolor{red}{'Euro'}} \\
\hline
\textbf{Error Explanation} & "EUR" is the correct entity for filtering. \\
\hline
\hline 
\textbf{Question} & List out the clients who choose statement of issuance after transaction are Disponent? \\
\hline
\hline 
\textbf{Gold SQL} &
\texttt{SELECT T2.client\_id \newline
FROM account AS T1 \newline
INNER JOIN disp AS T2 \newline
ON T1.account\_id = T2.account\_id \newline
WHERE T1.frequency = 'POPLATEK PO OBRATU' \newline
AND T2.type = \textcolor{blue}{'DISPONENT'}} \\
\hline
\textbf{Predicted SQL} &
\texttt{SELECT T2.client\_id \newline
FROM account AS T1 \newline
INNER JOIN disp AS T2 \newline
ON T1.account\_id = T2.account\_id \newline
WHERE T1.frequency = 'POPLATEK PO OBRATU' \newline
AND T2.type = \textcolor{red}{'Disponent'}} \\
\hline
\textbf{Error Explanation} & "DISPONENT" is the correct entity for filtering.  \\
\hline
\end{longtable}

\subsubsection{Revision: }
Description: The revision node incorrectly changes the correct predicted SQL into an incorrectly predicted SQL.
\begin{longtable}{p{0.2\textwidth} p{0.8\textwidth}}
\hline
\hline 
\textbf{Question} & In Los Angeles how many schools have more than 500 free meals but less than 700 free or reduced price meals for K-12? \\
\hline
\hline 
\textbf{Gold SQL} &
\texttt{SELECT COUNT(CDSCode) \newline
FROM frpm \newline
WHERE County Name = 'Los Angeles' \newline
AND Free Meal Count (K-12) > 500 \newline
AND \textcolor{blue}{'Free Meal Count (K-12)} < 700} \\
\hline
\textbf{Predicted SQL} &
\texttt{SELECT COUNT(CDSCode) \newline
FROM frpm \newline
WHERE Free Meal Count (K-12) > 500 \newline
AND \textcolor{red}{'FRPM Count (K-12)} < 700 \newline
AND County Name = 'Los Angeles'} \\
\hline
\textbf{Error Explanation} & "Free Meal Count (K-12)" is the correct column for filtering.  \\
\hline
\hline 
\textbf{Question} & Find and list the names of sets which doesn't have Japanese translation but have Korean translation. \\
\hline
\hline 
\textbf{Gold SQL} &
\texttt{SELECT name \newline
FROM sets \newline
WHERE \textcolor{blue}{code IN ( \newline
SELECT setCode \newline
FROM set\_translations \newline
WHERE language = 'Korean' \newline
AND language NOT LIKE 'Japanese') }} \\
\hline
\textbf{Predicted SQL} &
\texttt{SELECT T1.name \newline
FROM sets AS T1 \newline
WHERE \textcolor{red}{EXISTS (SELECT 1 \newline
FROM set\_translations AS T2 \newline
WHERE T1.code = T2.setCode \newline
AND T2.language = 'Korean') \newline
AND NOT EXISTS (SELECT 1 \newline
FROM set\_translations AS T3 \newline
WHERE T1.code = T3.setCode \newline
AND T3.language LIKE 'Japanese')}} \\
\hline
\textbf{Error Explanation} & Incorrect logic in subquery. \\
\hline
\end{longtable}

\subsubsection{Malformed Date: }
Description: The date format in the predicted SQL is incorrect.
\begin{longtable}{p{0.2\textwidth} p{0.8\textwidth}}
\hline
\hline 
\textbf{Question} & Which year recorded the most consumption of gas paid in CZK? \\
\hline
\hline 
\textbf{Evidence} & The first 4 strings of the values in the table yearmonth can represent year. \\
\hline
\textbf{Gold SQL} &
\texttt{SELECT \textcolor{blue}{SUBSTRING(T2.Date, 1, 4)} \newline
FROM customers AS T1 \newline
INNER JOIN yearmonth AS T2 \newline
ON T1.CustomerID = T2.CustomerID \newline
WHERE T1.Currency = 'CZK' \newline
GROUP BY \textcolor{blue}{SUBSTRING(T2.Date, 1, 4)} \newline
ORDER BY SUM(T2.Consumption) DESC \newline
LIMIT 1} \\
\hline
\textbf{Predicted SQL} &
\texttt{SELECT \textcolor{red}{strftime('\%Y', T2.Date)} AS Year \newline
FROM yearmonth T2 \newline
JOIN customers T1 \newline
ON T2.CustomerID = T1.CustomerID \newline
WHERE T1.Currency = 'CZK' \newline
GROUP BY \textcolor{red}{Year} \newline
ORDER BY SUM(T2.Consumption) DESC \newline
LIMIT 1} \\
\hline
\textbf{Error Explanation} & Incorrect year formating. \\
\hline
\end{longtable}

\subsubsection{Evidence: }
Description: The evidence was not used correctly in the predicted SQL.
\begin{longtable}{p{0.2\textwidth} p{0.8\textwidth}}
\hline
\hline 
\textbf{Question} & Name schools in Riverside which the average of average math score for SAT is greater than 400, what is the funding type of these schools? \\
\hline
\hline 
\textbf{Evidence} & Average of average math = sum(average math scores) / count(schools). \\
\hline
\textbf{Gold SQL} &
\texttt{SELECT T1.sname, T2.\`Charter Funding Type\` \newline
FROM satscores AS T1 \newline
INNER JOIN frpm AS T2 \newline
ON T1.cds = T2.CDSCode \newline
WHERE T2.District Name LIKE 'Riverside\%' \newline
GROUP BY T1.sname, T2.Charter Funding Type\` \newline
HAVING \textcolor{blue}{CAST(SUM(T1.AvgScrMath) AS REAL) / COUNT(T1.cds) > 400}} \\
\hline
\textbf{Predicted SQL} &
\texttt{SELECT DISTINCT T1.CDSCode, T1.FundingType \newline
FROM schools AS T1 \newline
INNER JOIN satscores AS T2 \newline
ON T1.CDSCode = T2.cds \newline
WHERE T1.County = 'Riverside' \newline
AND \textcolor{red}{T2.AvgScrMath > 400}} \\
\hline
\textbf{Error Explanation} & Did not use the equation in the evidence. \\
\hline
\hline 
\textbf{Question} & What is the detailed content of the comment of the post which was created on 7/19/2010 7:37:33 PM? \\
\hline
\hline 
\textbf{Evidence} & detailed content of the comment refers to Text; created on 7/19/2010 7:37:33 PM CreationDate = 2010-07-19 19:37:33.0 \\
\hline
\textbf{Gold SQL} &
\texttt{SELECT T1.Text \newline
FROM comments AS T1 \newline
INNER JOIN posts AS T2 \newline
ON T1.PostId = T2.Id \newline
WHERE T1.CreationDate = \textcolor{blue}{'2010-07-19 19:37:33.0'}} \\
\hline
\textbf{Predicted SQL} &
\texttt{SELECT T2.Text \newline
FROM posts AS T1 \newline
INNER JOIN comments AS T2 \newline
ON T1.Id = T2.PostId \newline
WHERE T1.CreationDate = \textcolor{red}{'2010-07-19 19:37:33'}} \\
\hline
\textbf{Error Explanation} & Did not use the time formatting in the evidence. \\
\hline
\end{longtable}

\subsubsection{NULL/DISTINCT/GROUP BY: }
Description: Missing NULL, DISTINCT, or GROUP BY in the predicted SQL.
\begin{longtable}{p{0.2\textwidth} p{0.8\textwidth}}
\hline
\hline 
\textbf{Question} & When was the oldest mythic card released and what are its legal play formats? \\
\hline
\hline 
\textbf{Evidence} & the oldest card refers to MIN(originalReleaseDate); mythic card refers to rarity = 'mythic'; legal play refers to status = 'legal'; play format refers to format \\
\hline
\textbf{Gold SQL} &
\texttt{SELECT T1.originalReleaseDate, T2.format \newline
FROM cards AS T1 \newline
INNER JOIN legalities AS T2 \newline
ON T1.uuid = T2.uuid \newline
WHERE T1.rarity = 'mythic' \newline
AND \textcolor{blue}{T1.originalReleaseDate IS NOT NULL} \newline
AND T2.status = 'Legal' \newline
ORDER BY T1.originalReleaseDate \newline
LIMIT 1} \\
\hline
\textbf{Predicted SQL} &
\texttt{SELECT T1.originalReleaseDate, T2.format \newline
FROM cards AS T1 \newline
INNER JOIN legalities AS T2 \newline
ON T1.uuid = T2.uuid \newline
WHERE T1.rarity = 'mythic' \newline
AND T2.status = 'Legal' \newline
ORDER BY T1.originalReleaseDate ASC \newline
LIMIT 1} \\
\hline
\textbf{Error Explanation} & Did not check if the selected column has NULL values. \\
\hline
\hline 
\textbf{Question} & Name the foreign name of the card that has abzan watermark? List out the type of this card. \\
\hline
\hline
\textbf{Evidence} &  \\
\hline
\textbf{Gold SQL} &
\texttt{SELECT \textcolor{blue}{DISTINCT} T1.name, T1.type \newline
FROM cards AS T1 \newline
INNER JOIN foreign\_data AS T2 \newline
ON T2.uuid = T1.uuid \newline
WHERE T1.watermark = 'abzan'} \\
\hline
\textbf{Predicted SQL} &
\texttt{SELECT T2.name, T2.type \newline
FROM cards AS T1 \newline
INNER JOIN foreign\_data AS T2 \newline
ON T1.uuid = T2.uuid \newline
WHERE T1.watermark = 'abzan'} \\
\hline
\textbf{Error Explanation} & Did not use the DISTINCT keyword for the selected columns. \\
\hline
\end{longtable}

\subsubsection{Operation:}
Description: Incorrect comparison or division operations in predicted SQL.
\begin{longtable}{p{0.2\textwidth} p{0.8\textwidth}}
\hline
\hline 
\textbf{Question} & Among the male patients who have a normal level of white blood cells, how many of them have an abnormal fibrinogen level? \\
\hline
\hline
\textbf{Evidence} & male patients refers to Sex = 'M'; normal level of white blood cells refers to WBC between 3.5 and 9.0; abnormal fibrinogen level refers to FG $<=$ 150 or FG $>=$ 450; Don't compute repetitive ones. \\
\hline
\textbf{Gold SQL} &
\texttt{SELECT COUNT(DISTINCT T1.ID) \newline
FROM Patient AS T1 \newline
INNER JOIN Laboratory AS T2 \newline
ON T1.ID = T2.ID \newline
WHERE \textcolor{blue}{T2.FG <= 150 OR T2.FG >= 450} \newline
AND T2.WBC > 3.5 \newline
AND T2.WBC < 9.0 \newline
AND T1.SEX = 'M'} \\
\hline
\textbf{Predicted SQL} &
\texttt{SELECT COUNT(DISTINCT T1.ID) \newline
FROM Patient AS T1 \newline
INNER JOIN Laboratory AS T2 \newline
ON T1.ID = T2.ID \newline
WHERE T1.SEX = 'M' \newline
AND T2.WBC BETWEEN 3.5 AND 9.0 \newline
AND \textcolor{red}{(T2.FG < 150 OR T2.FG > 450)}} \\
\hline
\textbf{Error Explanation} & Incorrect comparison operation. \\
\hline
\hline 
\textbf{Question} & Among all chemical compounds identified in the database, what percent of compounds form a triple-bond. \\
\hline
\hline
\textbf{Evidence} & triple bond refers to bond\_type = '\#'; \\
\hline
\textbf{Gold SQL} &
\texttt{SELECT \textcolor{blue}{CAST(COUNT(CASE WHEN T.bond\_type = '\#' THEN T.bond\_id ELSE NULL END) AS REAL) * 100 / COUNT(T.bond\_id)} \newline
FROM bond AS T} \\
\hline
\textbf{Predicted SQL} &
\texttt{SELECT \textcolor{red}{CAST(COUNT(DISTINCT T1.molecule\_id) AS REAL) * 100 \newline
/ (SELECT COUNT(\*) FROM molecule)} AS percentage \newline
FROM bond AS T1 \newline
WHERE T1.bond\_type = '\#'} \\
\hline
\textbf{Error Explanation} & Incorrect division operation with the wrong column. \\
\hline
\end{longtable}

\subsubsection{SELECT: }
Description: Incorrect formatting in SELECT in the predicted SQL.
\begin{longtable}{p{0.2\textwidth} p{0.8\textwidth}}
\hline
\hline 
\textbf{Question} & Provide the full name and email address of the Student\_Club's Secretary. \\
\hline
\hline
\textbf{Evidence} & full name refers to first\_name, last\_name; 'Secretary' is a position of Student Club \\
\hline
\textbf{Gold SQL} &
\texttt{SELECT first\_name, last\_name, email \newline
FROM member \newline
WHERE position = 'Secretary'} \\
\hline
\textbf{Predicted SQL} &
\texttt{SELECT first\_name \textcolor{red}{\textbar\textbar ' ' \textbar\textbar} last\_name AS full\_name, email \newline
FROM member \newline
WHERE position = 'Secretary'} \\
\hline
\textbf{Error Explanation} & Incorrect column formatting in SELECT. \\
\hline
\hline 
\end{longtable}

\subsection{Vague Question}
\subsubsection{Column: }
Description: The question and evidence do not specify which columns should be included in the SELECT statement, making it impossible to determine the correct columns from the provided information and database. This leads to incorrect or additional columns in the gold SQL.
\begin{longtable}{p{0.2\textwidth} p{0.8\textwidth}}
\hline
\hline 
\textbf{Question} & \textcolor{red}{What are the cards for set OGW? State the colour for these cards.} \\
\hline
\hline
\textbf{Evidence} & set OGW refers to setCode = 'OGW'; \\
\hline
\textbf{Gold SQL} &
\texttt{SELECT id, colors \newline
FROM cards \newline
WHERE id IN ( SELECT id \newline
FROM set\_translations \newline
WHERE setCode = 'OGW' )} \\
\hline
\textbf{Predicted SQL} &
\texttt{SELECT name, colors \newline
FROM cards \newline
WHERE setCode = 'OGW'} \\
\hline
\textbf{Error Explanation} & Unclear if the question is asking for name or id. \\
\hline
\hline 
\textbf{Question} & \textcolor{red}{Which of these players performs the best in crossing actions, Alexis, Ariel Borysiuk or Arouna Kone?} \\
\hline
\hline
\textbf{Evidence} & player who perform best in crossing actions refers to MAX(crossing); \\
\hline
\textbf{Gold SQL} &
\texttt{SELECT t1.player\_name, t2.crossing \newline
FROM Player AS t1 \newline
INNER JOIN Player\_Attributes AS t2 \newline
ON t1.player\_api\_id = t2.player\_api\_id \newline
WHERE t1.player\_name IN ('Alexis', 'Ariel Borysiuk', 'Arouna Kone') \newline
ORDER BY t2.crossing DESC \newline
LIMIT 1} \\
\hline
\textbf{Predicted SQL} &
\texttt{SELECT T1.player\_name \newline
FROM Player AS T1 \newline
INNER JOIN Player\_Attributes AS T2 \newline
ON T1.player\_fifa\_api\_id = T2.player\_fifa\_api\_id \newline
WHERE T1.player\_name IN ('Alexis', 'Ariel Borysiuk', 'Arouna Kone') \newline
ORDER BY T2.crossing DESC \newline
LIMIT 1} \\
\hline
\textbf{Error Explanation} & Unclear from the question if the output should include both the names and the crossing actions. \\
\hline
\end{longtable}

\subsection{Incorrect Golden SQL}
\subsubsection{Evidence: }
Description: The gold SQL does not use the evidence correctly.
\begin{longtable}{p{0.2\textwidth} p{0.8\textwidth}}
\hline
\hline 
\textbf{Question} & What is the percentage of Story Spotlight cards that also have a text box? List them by their ID. \\
\hline
\hline
\textbf{Evidence} & Story Spotlight cards that do not have a text box refers to isStorylight = 1 and isTextless = 1; Percentage refer to DIVIDE(SUM(count(id) where isStorylight = 1), SUM(count(id))) * 100 \\
\hline
\textbf{Gold SQL} &
\texttt{SELECT CAST(SUM(CASE WHEN isTextless = 0 THEN 1 ELSE 0 END) AS REAL) \* 100 / COUNT(id) \newline
FROM cards \newline
WHERE \textcolor{red}{isStorySpotlight = 1}} \\
\hline
\textbf{Predicted SQL} &
\texttt{SELECT CAST(SUM(CASE WHEN \textcolor{blue}{isStorySpotlight = 1} AND isTextless = 0 THEN 1 ELSE 0 END) AS REAL) \* 100 / COUNT(id) AS percentage \newline
FROM cards} \\
\hline
\textbf{Error Explanation} & The filtering should be done in the numerator as shown in the evidence. \\
\hline
\hline 
\textbf{Question} & Who is the illustrator that illustrated the least amount of cards? List the format of play of the cards that he/she illustrated. \\
\hline
\hline
\textbf{Evidence} & format of the cards refers to format; illustrator refers to artist; the least amount of cards refers to MIN(artist) \\
\hline
\textbf{Gold SQL} &
\texttt{SELECT T1.artist, T2.format \newline
FROM cards AS T1 \newline
INNER JOIN legalities AS T2 \newline
ON T2.uuid = T1.uuid \newline
GROUP BY T1.artist \newline
ORDER BY \textcolor{red}{COUNT(T1.id)} ASC \newline
LIMIT 1} \\
\hline
\textbf{Predicted SQL} &
\texttt{SELECT T1.artist, T2.format \newline
FROM cards AS T1 \newline
INNER JOIN legalities AS T2 \newline
ON T1.uuid = T2.uuid \newline
GROUP BY T1.artist, T2.format \newline
ORDER BY \textcolor{blue}{COUNT(T1.artist)} ASC \newline
LIMIT 1} \\
\hline
\textbf{Error Explanation} & The gold SQL did not follow the evidence instruction to use the right column in MIN(artist). \\
\hline
\end{longtable}

\subsubsection{Description: }
Description: The gold SQL does not utilize the column or value descriptions provided in the database, leading to incorrect SQL.
\begin{longtable}{p{0.2\textwidth} p{0.8\textwidth}}
\hline
\hline 
\textbf{Question} & Which gas station has the highest amount of revenue? \\
\hline
\hline
\textbf{Gold SQL} &
\texttt{SELECT GasStationID \newline
FROM transactions\_1k \newline
GROUP BY GasStationID \newline
ORDER BY \textcolor{red}{SUM(Price)} DESC \newline
LIMIT 1} \\
\hline
\textbf{Predicted SQL} &
\texttt{SELECT T2.GasStationID \newline
FROM gasstations AS T1 \newline
INNER JOIN transactions\_1k AS T2 \newline
ON T1.GasStationID = T2.GasStationID \newline
WHERE T2.Amount IS NOT NULL \newline
AND T2.Price IS NOT NULL \newline
GROUP BY T2.GasStationID \newline
ORDER BY \textcolor{blue}{SUM(T2.Amount * T2.Price)} DESC \newline
LIMIT 1} \\
\hline
\textbf{Error Explanation} & The description in the database indicates revenue = amount*price. \\
\hline
\hline 
\textbf{Question} & What's the finish time for the driver who ranked second in 2008's Australian Grand Prix? \\
\hline
\textbf{Gold SQL} &
\texttt{SELECT T1.time \newline
FROM results AS T1 \newline
INNER JOIN races AS T2 \newline
ON T1.raceId = T2.raceId \newline
WHERE \textcolor{red}{T1.rank = 2} \newline
AND T2.name = 'Australian Grand Prix' \newline
AND T2.year = 2008} \\
\hline
\textbf{Predicted SQL} &
\texttt{SELECT T2.time \newline
FROM races AS T1 \newline
INNER JOIN results AS T2 \newline
ON T1.raceId = T2.raceId \newline
WHERE T1.name = 'Australian Grand Prix' \newline
AND T1.year = 2008 \newline
AND \textcolor{blue}{T2.positionOrder = 2}} \\
\hline
\textbf{Error Explanation} & The description in the database indicates the position order is the finishing order of positions while rank is the starting rank positioned by fastest lap speed. \\
\hline
\end{longtable}

\subsubsection{Aggregation: }
Description: The gold SQL misses aggregation operations.
\begin{longtable}{p{0.2\textwidth} p{0.8\textwidth}}
\hline
\hline 
\textbf{Question} & What is the amount of debt that client number 992 has, and how is this client doing with payments? \\
\hline
\hline
\textbf{Gold SQL} &
\texttt{SELECT \textcolor{red}{T3.amount}, T3.status \newline
FROM client AS T1 \newline
INNER JOIN account AS T2 \newline
ON T1.district\_id = T2.district\_id \newline
INNER JOIN loan AS T3 \newline
ON T2.account\_id = T3.account\_id \newline
WHERE T1.client\_id = 992} \\
\hline
\textbf{Predicted SQL} &
\texttt{SELECT \textcolor{blue}{SUM(T3.amount)} AS total\_debt, T3.status \newline
FROM client AS T1 \newline
INNER JOIN disp AS T2 \newline
ON T1.client\_id = T2.client\_id \newline
INNER JOIN loan AS T3 \newline
ON T2.account\_id = T3.account\_id \newline
WHERE T1.client\_id = 992 \newline
GROUP BY T3.status} \\
\hline
\textbf{Error Explanation} & Gold SQL did not use SUM to sum the full amount. \\
\hline
\hline 
\textbf{Question} & How many cards have frame effect as extendedart? List out the id of those cards. \\
\hline
\hline
\textbf{Gold SQL} &
\texttt{SELECT \textcolor{red}{id} \newline
FROM cards \newline
WHERE frameEffects = 'extendedart' \newline
GROUP BY id} \\
\hline
\textbf{Predicted SQL} &
\texttt{SELECT \textcolor{blue}{COUNT(id)} \newline
FROM cards \newline
WHERE frameEffects = 'extendedart' \newline
UNION ALL \newline
SELECT id \newline
FROM cards \newline
WHERE frameEffects = 'extendedart'} \\
\hline
\textbf{Error Explanation} & The gold SQL did not return a count as asked by the question. \\
\hline
\end{longtable}

\subsubsection{Column: }
Description: The gold SQL does not have all the expected columns from the question in SELECT.
\begin{longtable}{p{0.2\textwidth} p{0.8\textwidth}}
\hline
\hline 
\textbf{Question} & List out all post that are related to post ID 61217 and what is the popularity of this post? \\
\hline
\hline
\textbf{Evidence} & post related refers to RelatedPostId; popularity refers to ViewCount \\
\hline
\textbf{Gold SQL} &
\texttt{SELECT T1.ViewCount \newline
FROM posts AS T1 \newline
INNER JOIN postLinks AS T2 \newline
ON T1.Id = T2.PostId \newline
WHERE T2.PostId = 61217} \\
\hline
\textbf{Predicted SQL} &
\texttt{SELECT \textcolor{blue}{T2.RelatedPostId}, T1.ViewCount \newline
FROM posts AS T1 \newline
INNER JOIN postLinks AS T2 \newline
ON T1.Id = T2.PostId \newline
WHERE T2.PostId = 61217} \\
\hline
\textbf{Error Explanation} & The gold SQL did not select all the columns asked by the question. \\
\hline
\hline 
\textbf{Question} & List down the bond type for molecules from molecule id TR000 to TR050. \\
\hline
\hline
\textbf{Evidence} & double bond refers to bond\_type = ' = '; single bond refers to bond\_type = '-'; triple bond refers to bond\_type = '\#'; \\
\hline
\textbf{Gold SQL} &
\texttt{SELECT \textcolor{red}{T2.molecule\_id}, T2.bond\_type \newline
FROM molecule AS T1 \newline
INNER JOIN bond AS T2 \newline
ON T1.molecule\_id = T2.molecule\_id \newline
WHERE T1.molecule\_id BETWEEN 'TR000' AND 'TR050'} \\
\hline
\textbf{Predicted SQL} &
\texttt{SELECT T2.bond\_type \newline
FROM molecule AS T1 \newline
INNER JOIN bond AS T2 \newline
ON T1.molecule\_id = T2.molecule\_id \newline
WHERE T1.molecule\_id LIKE 'TR0\%' \newline
AND T1.molecule\_id BETWEEN 'TR000' AND 'TR050'} \\
\hline
\textbf{Error Explanation} & The gold SQL has more columns selected than what the question asked for. \\
\hline
\end{longtable}

\subsubsection{Filtering: }
Description: The gold SQL does not filter using the correct entity or is missing necessary filters.
\begin{longtable}{p{0.2\textwidth} p{0.8\textwidth}}
\hline
\hline 
\textbf{Question} & How many male customers who are living in North Bohemia have average salary greater than 8000? \\
\hline
\hline
\textbf{Evidence} & Male means that gender = 'M'; A3 refers to region; A11 pertains to average salary. \\
\hline
\textbf{Gold SQL} &
\texttt{SELECT COUNT(T1.client\_id) \newline
FROM client AS T1 \newline
INNER JOIN district AS T2 \newline
ON T1.district\_id = T2.district\_id \newline
WHERE T1.gender = 'M' \newline
AND T2.A3 = '\textcolor{red}{North} Bohemia' \newline
AND T2.A11 > 8000} \\
\hline
\textbf{Predicted SQL} &
\texttt{SELECT COUNT(T1.client\_id) \newline
FROM client AS T1 \newline
INNER JOIN district AS T2 \newline
ON T1.district\_id = T2.district\_id \newline
WHERE T1.gender = 'M' \newline
AND T2.A3 = '\textcolor{blue}{north} Bohemia' \newline
AND T2.A11 > 8000} \\
\hline
\textbf{Error Explanation} & The correct entity name has a lower case in north. \\
\hline
\hline 
\textbf{Question} & How many users from New York have a teacher and supporter badge? \\
\hline
\hline
\textbf{Evidence} & "Supporter" and "Teachers" are both Name of badge; 'New York' is the Location; user refers to UserId \\
\hline
\textbf{Gold SQL} &
\texttt{SELECT COUNT(DISTINCT T1.Id) \newline
FROM badges AS T1 \newline
INNER JOIN users AS T2 \newline
ON T1.UserId = T2.Id \newline
WHERE T1.Name IN ('Supporter', 'Teacher') \newline
AND T2.Location = 'New York'} \\
\hline
\textbf{Predicted SQL} &
\texttt{SELECT COUNT(DISTINCT T1.Id) \newline
FROM users AS T1 \newline
INNER JOIN badges AS T2 \newline
ON T1.Id = T2.UserId \newline
WHERE T1.Location = 'New York' \newline
AND T2.Name IN ('Supporter', 'Teacher') \newline
GROUP BY T1.Id \newline
\textcolor{blue}{HAVING COUNT(DISTINCT T2.Name) = 2}} \\
\hline
\textbf{Error Explanation} & The gold SQL missed a filtering that indicates a user is both teacher and a supporter. \\
\hline
\end{longtable}

\subsubsection{NULL/DISTINCT/GROUP BY: }
Description: The gold SQL missed checking NULL, adding DISTINCT, or adding GROUP BY.
\begin{longtable}{p{0.2\textwidth} p{0.8\textwidth}}
\hline
\hline 
\textbf{Question} & Calculate the total amount spent on speaker gifts and list the name of the event they were spent on. \\
\hline
\hline
\textbf{Evidence} & total amount spent = SUM(spent) where category = 'Speaker Gifts' \\
\hline
\textbf{Gold SQL} &
\texttt{SELECT SUM(T1.spent), T2.event\_name \newline
FROM budget AS T1 \newline
INNER JOIN event AS T2 \newline
ON T1.link\_to\_event = T2.event\_id \newline
WHERE T1.category = 'Speaker Gifts'} \\
\hline
\textbf{Predicted SQL} &
\texttt{SELECT SUM(T2.spent), T1.event\_name \newline
FROM event AS T1 \newline
INNER JOIN budget AS T2 \newline
ON T1.event\_id = T2.link\_to\_event \newline
WHERE T2.category = 'Speaker Gifts' \newline
\textcolor{blue}{GROUP BY T1.event\_name}} \\
\hline
\textbf{Error Explanation} & The gold SQL misses GROUP BY. \\
\hline
\end{longtable}